\documentclass[sigconf]{acmart}
\acmSubmissionID{2120}
\renewcommand\footnotetextcopyrightpermission[1]{}
\usepackage{bigstrut}
\usepackage{graphicx}
\usepackage{pifont}
\usepackage{float}

\begin{document}

\title{Beyond Words: AuralLLM and SignMST-C for Sign Language Production and Bidirectional Accessibility}






\author{
    \textbf{
        Yulong Li\textsuperscript{1,2*}, 
        Yuxuan Zhang\textsuperscript{1*}, 
        Feilong Tang\textsuperscript{2,3}, 
        Ming Hu\textsuperscript{2,3}, 
        Zhixiang Lu\textsuperscript{1}, 
        Haochen Xue\textsuperscript{1}, 
        Jianghao Wu\textsuperscript{2}, 
        Mian Zhou\textsuperscript{1}, 
        Kang Dang\textsuperscript{1},
        Chong Li\textsuperscript{1},
        Yifang Wang\textsuperscript{1}, 
        Imran Razzak\textsuperscript{2\textdagger}, 
        Jionglong Su\textsuperscript{1\textdagger}     
    }\\
    \hspace*{-2em} 
    \fontsize{9.5pt}{11.6pt}\selectfont{\textsuperscript{1} School of Artificial Intelligence and Advanced Computing, Xi'an Jiaotong-Liverpool University} \\
    \fontsize{9.5pt}{11.6pt}\selectfont{\textsuperscript{2} Mohamed bin Zayed University of Artificial Intelligence} \\
    \fontsize{9.5pt}{11.6pt}\selectfont{\textsuperscript{3} Monash University} \\
    \vspace{-0.1em}
    Imran.Razzak@mbzuai.ac.ae, Jionglong.Su@xjtlu.edu.cn
}

\renewcommand{\shortauthors}{Yulong et al.}



\begin{abstract}
Sign language is the primary communication mode for 72 million hearing-impaired individuals worldwide, necessitating effective bidirectional Sign Language Production and Sign Language Translation systems. However, functional bidirectional systems require a unified linguistic environment, hindered by the lack of suitable unified datasets, particularly those providing the necessary pose information for accurate Sign Language Production (SLP) evaluation. Concurrently, current SLP evaluation methods like back-translation ignore pose accuracy, and high-quality coordinated generation remains challenging. To create this crucial environment and overcome these challenges, we introduce CNText2Sign and CNSign, which together constitute the first unified dataset aimed at supporting bidirectional accessibility systems for Chinese sign language; CNText2Sign provides 15,000 natural language-to-sign mappings and standardized skeletal keypoints for 8,643 vocabulary items supporting pose assessment. Building upon this foundation, we propose the AuraLLM model, which leverages a decoupled architecture with CNText2Sign's pose data for novel direct gesture accuracy assessment. The model employs retrieval augmentation and Cascading Vocabulary Resolution to handle semantic mapping and out-of-vocabulary words, and achieves all-scenario production with controllable coordination of gestures and facial expressions via pose-conditioned video synthesis. Concurrently, our Sign Language Translation model SignMST-C employs targeted self-supervised pretraining for dynamic feature capture, achieving new SOTA results on PHOENIX2014-T with BLEU-4 scores up to 32.08. AuraLLM establishes a strong performance baseline on CNText2Sign with a BLEU-4 score of 50.41 under direct evaluation.


\vspace{-1em}
\end{abstract}


\begin{teaserfigure}
\centering
  \includegraphics[width=0.92\textwidth]{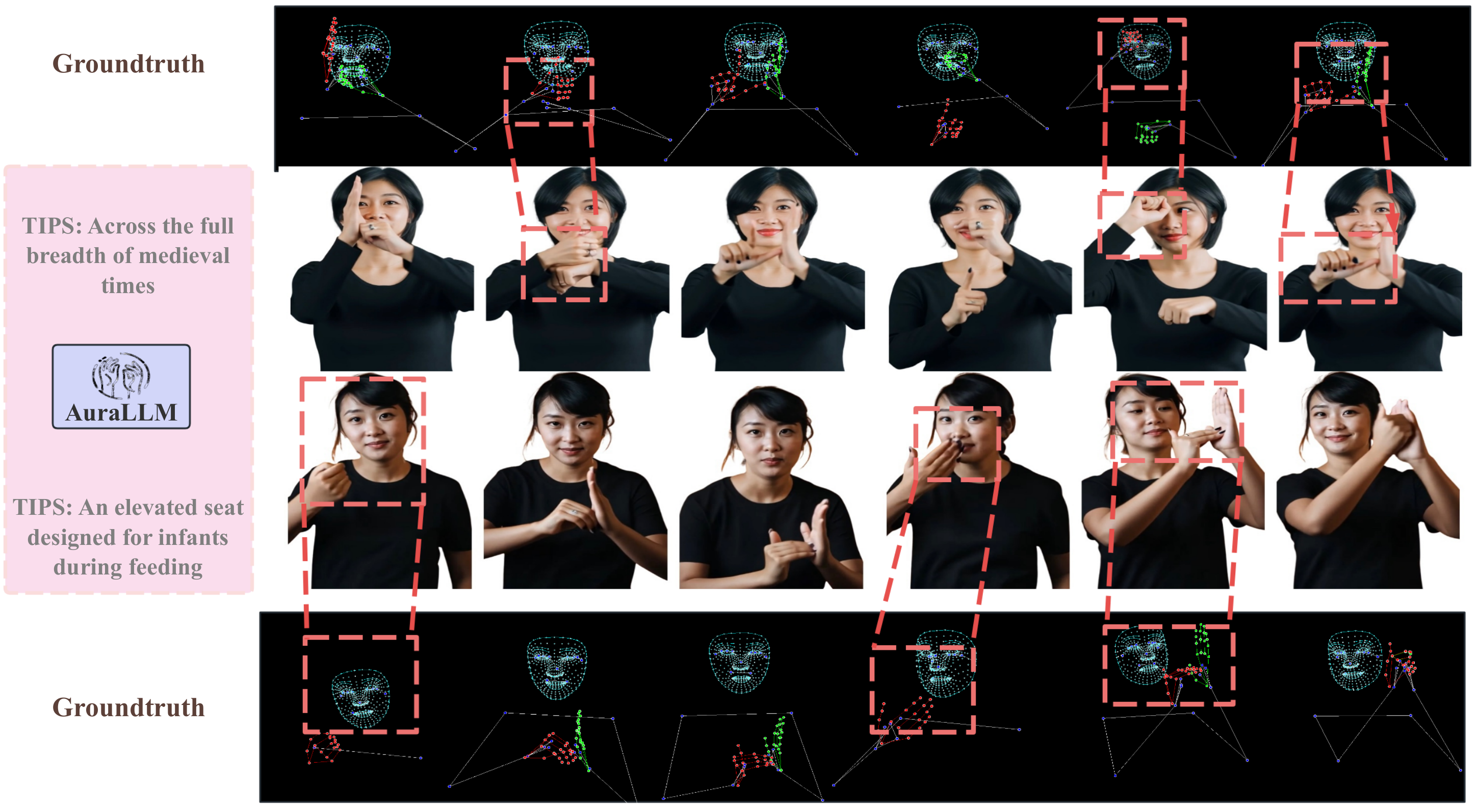}
  \vspace{-1em}
  \caption{AuraLLM generating sign language video frames with coordinated facial expressions and hand gestures from natural language, compared against ground truth skeletal representations (key areas highlighted)}
  \label{fig1show}
\end{teaserfigure}
\maketitle

\vspace{-1em}
\section{Introduction}
\renewcommand{\thefootnote}{\arabic{footnote}}

\begin{figure}[t!] 
  \centering
  \includegraphics[width=0.5\textwidth]{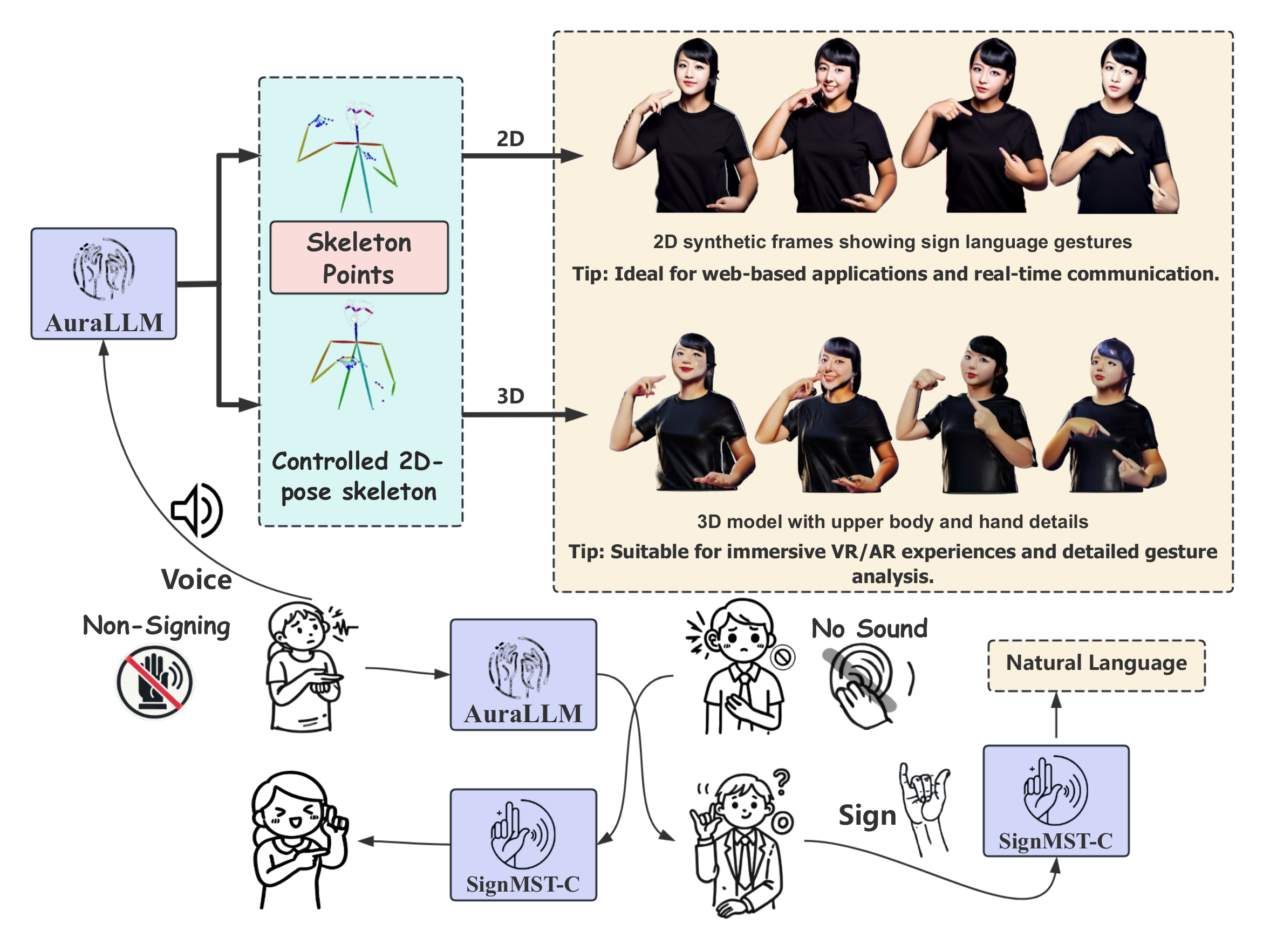}
  \vspace{-2em}
  \caption{BeyondWords: Enabling Barrier-Free Communication Between Hearing and Hearing-Impaired Individuals.}

  \label{BeyondWords}
  \vspace{-2.5em}
\end{figure}

Approximately 72 million hearing-impaired individuals worldwide rely primarily on sign language for daily communication~\cite{secretariat2009, un_wfd_2023}. Achieving truly barrier-free, fluent bidirectional communication between the Deaf community and hearing individuals, as illustrated in Figure ~\ref{BeyondWords}, is a crucial goal for technological development. Although significant progress has been made individually in Sign Language Translation (SLT) and Sign Language Production (SLP) technologies~\cite{Chen2022TwoStreamNetworkForSignLanguage,Papastratis}, building a complete communication loop still faces substantial challenges~\cite{Bragg2019SignLanguageRecognition}. This primarily stems from two major bottlenecks: First, the high-precision requirement and implementation difficulty of SLP. As a visual language, sign language primarily expresses complex meanings through the precise coordination of hand movements, facial expressions, and body posture, a challenge addressed by the generation capabilities demonstrated in our work (illustrated in Figure ~\ref{fig1show}); Its inherent characteristics of spatiality and simultaneity in conveying information mean that even any subtle deviation in action form, spatial position, or accompanying non-manual features can potentially lead to significant and sometimes unintended differences in the conveyed semantics. Second, there is a notable lack of comprehensive datasets providing a unified and consistent data foundation for the two closely related and complementary tasks of Sign Language Translation (SLT) and Sign Language Production (SLP) within the same linguistic and cultural context. The absence of this foundational resource makes it extremely difficult to develop truly coherent, robust, and synergistic bidirectional systems. These two core and interdependent challenges jointly impede the emergence and practical implementation of truly functional bidirectional systems.

The high-level obstacles to integrated bidirectional communication manifest as persistent technical challenges across SLT and SLP, complicating the creation of synergistic systems:
\textit{(i)}, \textbf{SLT Processing Challenges:} Accurately translating sign language from video  requires overcoming hurdles in capturing rapid spatiotemporal motion~\cite{Zhang2023ControlNet} and understanding unique visual-grammatical structures ~\cite{camgoz2020signlanguagetransformersjoint,liang2024llavasltvisuallanguagetuning}. \textit{(ii)}, \textbf{SLP Generation Quality:} Generating high-quality sign language to video involves difficulties in producing coherent, natural motion ~\cite{duarte2021how2signlargescalemultimodaldataset,Huang2021FastAndHighQualitySignLanguage} and ensuring physical accuracy of manual actions and non-manual features like facial expressions. \textit{(iii)}, \textbf{SLP Mapping and Evaluation Challenges:} Effectively mapping text to sign requires resolving extensive Out-of-Vocabulary (OOV) issues due to vocabulary disparity. Moreover, reliance on back-translation as the dominant SLP evaluation metric is a fundamental barrier. Focusing on semantic congruence via text prevents assessment of crucial physical quality and pose accuracy, making it inadequate for validating high-fidelity, nuanced sign language production. \textit{(iv)}, \textbf{Data Scarcity and Limitations:} The field is constrained by scarce, costly, and limited annotated data. Existing datasets often lack sufficient scale, diversity, annotation granularity, particularly standardized pose data, for robust models. Critically, there is a profound lack of unified datasets explicitly designed for the parallel training and evaluation needed for integrated bidirectional systems.

These persistent challenges across SLT and SLP underscore the limitations of isolated development and highlight the critical need for seamless integration to realize true bidirectional accessibility. This integration necessitates operating within a unified linguistic context, which remains elusive largely due to critical limitations in existing sign language datasets. These limitations, such as the lack of unified structure needed for joint modeling, like comprehensive mappings between sign vocabulary and precise pose representations alongside aligned video-text data ~\cite{duarte2021how2signlargescalemultimodaldataset}, and the confinement of many datasets to limited domains or scenarios, collectively prevent the coherent development required for real-world, all-scenario communication and robust generalization. Consequently, establishing a data foundation that is both unified and comprehensive emerges as the crucial prerequisite for building the next generation of functional bidirectional accessibility systems.

To provide the needed unified, comprehensive data foundation, we introduce two CSL datasets: CNText2Sign and CNSign. CNText2Sign bridges natural language with sign vocabulary and precise pose representations for SLP. It provides 15,000 professionally annotated natural language to sign language vocabulary mappings for 8,643 core CSL items. For each vocabulary item, corresponding videos yield standardized pose sequences skeletal keypoints extracted via OpenPose and MediaPipe. This provides the missing vocabulary to pose linkage, a vital part of the unified structure for high fidelity SLP and direct pose evaluation. Its comprehensive coverage suits full scenario SLP, addressing prior dataset limitations. CNSign complements this with video to text mappings for SLT, using authentic context data to address scenario limitations. Together, these datasets furnish the essential unified CSL resources for advancing integrated bidirectional accessibility systems.


Building upon the unified and comprehensive data foundation provided by CNText2Sign and CNSign, we propose a dual-model architecture of Articulated Sign Language Production LLM (AuraLLM) and Sign Model Self-Translation with Correction (SignMST-C), designed to establish a complete bidirectional sign language accessibility system and achieve all-scenario sign language production. Specifically, AuraLLM addresses the back-translation evaluation limitation by decoupling SLP into semantic-to-representation translation and representation-to-video synthesis stages. The first stage converts natural language into intermediate sign language representations (standardized symbol sequences and their mapped skeletal poses), whose translation quality can be directly quantified using standard metrics like BLEU-n and ROUGE before video synthesis, ensuring semantic accuracy and avoiding the information loss and pose neglect issues of back-translation. To address the differences between natural language and sign language vocabularies, a Cascading Vocabulary Resolution (CVR) framework is adopted to process OOV vocabulary. In the video synthesis stage, ControlNet is first utilized to generate videos based on skeletal sequences to ensure accurate spatial pose execution, followed by Gen-3 Alpha optimization of critical hand and facial details, producing high-fidelity and naturally expressive videos. SignMST-C, targeting SLT tasks, designs a self-supervised pretraining method specifically for fast-motion video semantic reconstruction, enhancing feature learning capability for gestural dynamic regions through weighted perturbation processing on landmark data. The model employs various distillation losses to guide learning, ensuring temporal and semantic consistency between features of different modalities, while integrating a text correction network to improve translation accuracy for complex syntax. This integrated architecture, as clearly depicted in Figure 2, effectively lays the foundation for achieving coherent, fully complete bidirectional sign language communication.

The key contributions of this paper are three-fold:

\text{1)} We introduce the CNText2Sign and CNSign datasets, establishing a unified and reliable data foundation that supports the development of all-scenario Chinese Sign Language production and robust bidirectional accessibility systems. 

\text{2)} For SLT, we propose SignMST-C model. Through multimodal feature fusion and self-supervised pretraining method enhancing dynamic feature learning, complemented by a text correction network, SignMST-C achieves SOTA results on the Phoenix2014-T.

\text{3)} We propose AuraLLM, achieving all-scenario Sign Language Production with controllable coordination of gestures and facial expressions. AuraLLM's decoupled architecture and use of the CNText2Sign dataset establish a novel SLP evaluation paradigm allowing direct gesture accuracy assessment, resolving the core issues of information loss and lack of fidelity assessment in back-translation.

\vspace{-1em}
\section{Related Work}
\label{sec:related_work}

\textbf{Sign Language Datasets.} In recent years, research on SLP~\cite{Cox2002TESSA, Karpouzis2007GreekSignLanguage, Mazumder2021TranslatingSignLanguage, McDonald2016AutomatedTechnique, Segouat2009SignLanguageCoarticulation} has gradually shifted from traditional animation synthesis methods to data-driven approaches based on deep learning. However, its development is limited by lack of a unified data foundation for synergistic SLT and SLP, particularly in non-English sign languages. Existing datasets are mostly focused on a few languages, such as American Sign Language (ASL) and German Sign Language (GSL), and often restricted to specific topics, such as weather and daily phrases, which limits their applicability to a broader range of contexts~\cite{Saunders2020ProgressiveTransformers, Saunders2021Continuous3DSignLanguage, Saunders2021MixedSIGNals, Saunders2021SkeletalGraphSelfAttention}. Furthermore, the common lack of standardized skeletal pose information hinders both high-fidelity SLP generation and the direct physical accuracy evaluation needed to overcome back-translation limitations. Most datasets provide videos of isolated vocabulary or short phrases, lacking complex syntax and diverse contexts, making it challenging for generative models to produce natural and coherent sign language expressions in more varied situations~\cite{Duarte2021How2Sign}. Moreover, the commonly used back-translation method to evaluate sign language production has its limitations, as it struggles to capture the nuanced semantics and grammatical features of sign language~\cite{Bohacek2022SignPoseTransformer, Camgoz2018NeuralSignLanguageTranslation, Camgoz2020SignLanguageTransformers, Ko2019NeuralSignLanguageTranslation}. This issue is particularly pronounced in the domain of Chinese Sign Language (CSL), where existing datasets mostly consist of isolated words or basic phrases, lacking support for complex sentences and diverse scenes \cite{fang2024signllmsignlanguagesproduction}. To address these limitations, we introduce CNText2Sign, which surpasses isolated word/context constraints via multi-modal annotations and a large vocabulary. Its core contribution, explicit vocabulary-to-pose mappings with standardized sequences, is crucial for enabling direct SLP evaluation and advancing a unified data foundation for bidirectional systems.

\textbf{Large Language Model in SLP.}
\noindent\hspace{1em}
Large Language Models have demonstrated significant potential in SLP tasks 
due to their powerful natural language processing capabilities~\cite{Brown2020LanguageModels, Chowdhery2022Palm, Shanahan2022Talking, Taylor2022Galactica, Touvron2023Llama}. However, sign language, as a visual language, exhibits structural characteristics that differ significantly from the textual information processed by language models, presenting challenges in applying LLMs to the SLP domain. Current research typically applies LLMs by converting visual features of sign language, such as keypoints or skeletal poses, into sequential data for text-based pose generation, resembling a language modeling approach~\cite{Kreutzer2019JoeyNMT, Paszke2017AutoDiffPyTorch, Raffel2020ExploringTransferLearning, Raffel2020ExploringTransferLearning}. Since the vocabulary of sign language is much smaller than that of natural languages, LLMs often simplify details and lose information during the generation process, particularly in terms of emotions, intonation, and complex syntax, which lack direct counterparts in sign language. Some studies have attempted to combine LLMs with multimodal models by incorporating pose keypoints or gloss vocabulary as intermediate layers. However, the generation results still heavily depend on the LLM, and the design of prompts greatly impacts the output quality \cite{Brown2020LanguageModels, Chowdhery2022Palm, Radford2019LanguageModels}. Furthermore, sentences generated by LLMs often retain the syntax of natural language, which is not suitable for the concise expression style of sign language. And evaluating the quality of sign language generated by these LLM-based approaches typically relies on back-translation, which, as noted earlier, fundamentally fails to assess the physical fidelity and pose accuracy critical for fluent sign communication.
\begin{table*}[t]
  \centering
  \scalebox{0.83}{
  \begin{tabular}{l c c c c c c c c c}
  \hline
  \textbf{Name} & \textbf{Language} & \textbf{Vocab.} & \textbf{Duration} & \textbf{Avg Length} & 
  \textbf{Transcription} & \textbf{Gloss} & \textbf{Video Pose} & \textbf{Gloss Pose} & \textbf{Depth} \\ \hline
  SIGNUM~\cite{vonagris2010signum} & GSL & 450 & 55 & 25 & \ding{55} & \checkmark & \checkmark & \ding{55} & \ding{55} \\ 
  RWTH-Phoenix-2014T~\cite{Camgoz2018NeuralSignLanguageTranslation} & GSL & 3k & 11 & 9 & \ding{55} & \checkmark & \checkmark & \checkmark & \ding{55} \\ 
  Public DGS Corpus~\cite{hanke2020extending} & GSL & -- & 50 & 327 & \ding{55} & \checkmark & \checkmark & \checkmark & \ding{55} \\ 
  BSL Corpus~\cite{schembri2013building} & BSL & 5k & -- & 249 & \ding{55} & \checkmark & \checkmark & \checkmark & \checkmark \\ 
  NCSLGR~\cite{neidle2012new} & ASL & 1.8k & 5.3 & 4 & \ding{55} & \checkmark & \checkmark & \ding{55} & \ding{55} \\ 
  How2Sign~\cite{duarte2021how2signlargescalemultimodaldataset} & ASL & 16k & 79 & 11 & \ding{55} & \checkmark & \checkmark & \checkmark & \checkmark \\ 
  Prompt2Sign~\cite{fang2024signllmsignlanguagesproduction} & Multilingual & 40k & 200 & 40 & \checkmark & \checkmark & \checkmark & \ding{55} & \checkmark \\
  \hline
  Video-Based CSL~\cite{huang2018video} & CSL & 178 & 100 & 50 & \ding{55} & \checkmark & \ding{55} & \checkmark & \ding{55} \\
  \textbf{CNSign (ours)} & \textbf{CSL} & \textbf{34k} & \textbf{41.7} & \textbf{13.7} & \checkmark & \checkmark & \checkmark & \checkmark & \checkmark \\ \hline
  \end{tabular}
  }
  \caption{Detailed comparison of the proposed CNSign dataset with representative existing sign language datasets regarding language, scale, context, and key annotation features. Vocab size is in k=1000 units, Duration is in hours.}
  \label{tab:table1}
  \vspace{-2.5em}
\end{table*}

\textbf{SLT and Advances in Gloss-Free Approaches.}
\noindent\hspace{1em}
In recent years, to improve the effectiveness of SLT, many studies adopted sign gloss as an intermediate representation layer. SLRT~\cite{Camgoz2020SignLanguageTransformers} first introduced a Transformer-based encoder-decoder structure combined with Connectionist Temporal Classification (CTC) loss to align sign language representations with gloss sequences, significantly enhancing translation performance. However, the process of obtaining gloss annotations is complex and costly. To address this, STMC-T~\cite{Zhou2021SpatialTemporalMultiCueNetwork} proposed multi-stream learning, using both single-stream and cross-stream CTC losses to model sequence information more accurately, although this method has limited performance when dealing with complex syntax. SignBack~\cite{Zhou2021ImprovingSignLanguageTranslation} introduced back-translation techniques to improve the linguistic expressiveness of SLT, but back-translation struggles to ensure word-level precision, particularly in nuanced semantic expressions. Chen et al.~\cite{Chen2022SimpleMultiModalityTransferLearning, Chen2022TwoStreamNetworkForSignLanguage} explored the application of LLMs in SLT, leveraging the powerful generative capabilities of LLMs to enhance translation accuracy. However, LLMs still face challenges in aligning visual and linguistic modalities. Recently, gloss-free SLT has emerged as a new approach, with NSLT~\cite{Camgoz2018NeuralSignLanguageTranslation} utilizing a CNN+RNN architecture for end-to-end SLT. Despite its advancements, it still faces limitations in semantic and motion alignment due to the absence of intermediate representations. TSPNet~\cite{Li2020TSPNet} introduced a cross-scale attention mechanism to strengthen the capture of visual features, while CSGCR~\cite{Zhao2021ConditionalSentenceGeneration} improved SLT accuracy and fluency through a multi-module design. Nevertheless, challenges in cross-modal alignment remain unresolved. To address these issues, we propose an end-to-end SLT model combining self-supervised pretraining.

\vspace{-0.5em}
\section{Dataset Construction}
\label{thi:benchmark}

\noindent\textbf{Chinese Natural Language to Sign Language Gloss Mapping Dataset (CNText2Sign)}

To provide crucial resources for pose-aware Chinese SLP and enable direct evaluation, the CNText2Sign dataset features 15,000 mappings from Chinese natural language to sign language gloss. These mappings were professionally annotated; consistency was rigorously assessed on a 15\% sample, where each item was independently annotated by two annotators selected from a pool of six, achieving a high Fleiss' Kappa of 0.96, demonstrating strong annotation reliability. The dataset encompasses 8,643 basic CSL vocabulary items, each with corresponding video recordings. These videos were processed using both OpenPose~\cite{Cao2017OpenPose} and MediaPipe~\cite{lugaresi2019mediapipeframeworkbuildingperception} to extract standardized skeletal keypoints, providing complete pose sequences derived from both frameworks. This multimodal data, particularly the linked vocabulary-pose sequences, offers a standardized foundation for training SLP models and critically enables direct, quantitative evaluation of generated physical execution accuracy, advancing beyond semantic-only metrics.

\noindent\textbf{Chinese Sign Language to Natural Language Dataset (CNSign)} 

To address the scarcity of CSL data suitable for training robust SLT models capable of handling complex, contextual language, we introduce CNSign. Complementing CNText2Sign, CNSign provides the translation counterpart needed for developing integrated bidirectional CSL systems. This dataset contains 10,943 entries translating CSL videos into natural language. The video samples, primarily sourced from the Chinese news program "Focus On," feature realistic contexts and complex syntax, with durations ranging from 5 to 32 seconds. To handle the typical broadcast translation lag and ensure high accuracy and coherence, alignment between video and text was meticulously performed manually by seven trained experts. To rigorously assess alignment consistency, a 15\% sample was independently processed by two experts from the team of seven, yielding a Fleiss' Kappa of 0.93, indicating high reliability. Consequently, CNSign offers annotated data crucial for training and evaluating the capability of SLT models to handle real-world scenarios characterized by rich context and complex syntax, proving valuable for enhancing performance on complex linguistic expressions. 

\noindent\textbf{Dataset Splits} 

For standardized benchmarking, we provide predefined data splits for both the CNText2Sign and CNSign datasets. Each dataset was randomly partitioned at the entry level into training (80\%), validation (10\%), and test (10\%) sets. These standard splits are specifically provided to facilitate reproducible experiments and fair comparison of models developed using these resources. 

\noindent\textbf{Comparison with Existing Datasets} 

Table ~\ref{tab:table1} compares our CNSign dataset with representative resources, highlighting its provision of large-scale, contextual CSL data for translation tasks, addressing the domain scope and complexity limitations common in prior work. Although not detailed in the table, CNText2Sign complements this with another key contribution: it provides, to our knowledge, the first large-scale linkage between CSL vocabulary items and corresponding standardized pose sequences derived from OpenPose~\cite{Cao2017OpenPose} and MediaPipe~\cite{lugaresi2019mediapipeframeworkbuildingperception}. This vocabulary-pose association, absent in most existing datasets, is crucial for pose-aware SLP research and direct physical accuracy evaluation. Together, CNText2Sign and CNSign establish a unique and complementary foundation for bidirectional CSL accessibility research.
\vspace{-0.8em}

\begin{figure*}[t!] 
    \centering
    \includegraphics[width=0.93\textwidth]{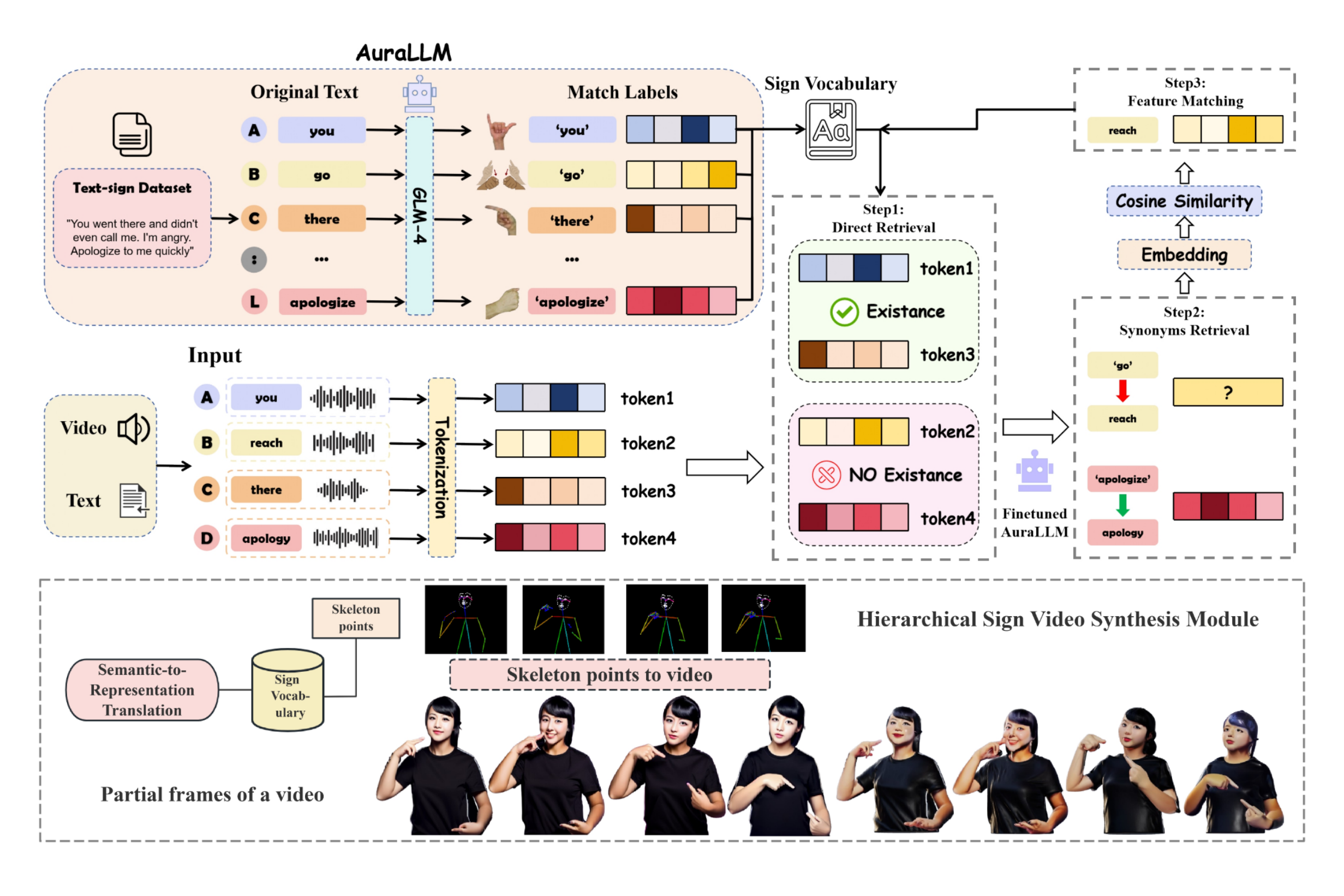}
    \vspace{-1cm}
    \caption{AuraLLM architecture overview. The Semantic-to-Representation Translation stage (upper section) converts natural language input to skeletal poses using an LLM and multi-level matching. These poses are then rendered into sign language video by the Hierarchical Sign Video Synthesis Module (lower section).}
    \vspace{-1.5em}
    \label{fig:txt2sign}
\end{figure*}

\section{Methodology}
\label{for:methodology}

\subsection*{SLP Framework}
The AuraLLM framework pioneers a decoupled approach to Sign Language Production (SLP), meticulously separating the process into two core stages: (1) Semantic-to-Representation Translation, converting natural language input into a pose-enriched intermediate sign representation, and (2) the Hierarchical Sign Video Synthesis Module (HSVSM), which renders this representation into a high-fidelity sign language video. This architecture, depicted in Figure ~\ref{fig:txt2sign}, is designed to leverage the strengths of large language models while ensuring precise control over the generated gestures and expressions, enabling direct evaluation of pose accuracy.
\vspace{-0.5em}
\subsection*{Semantic-to-Representation Translation Module} 

This module translates Natural Language (NL) input into a pose-enriched intermediate sign representation $\{(s_i, P_i)\}$. Its core is a fine-tuned Large Language Model (LLM) that leverages the CNText2Sign dataset to learn the mapping from NL to CSL symbols $s$ and intrinsically associates each symbol $s$ with its corresponding standard skeletal pose sequence $P$.

The translation process begins by tokenizing the input NL sequence, which is then processed by the fine-tuned LLM to understand contextual semantics. To accurately map the input to the appropriate CSL symbol-pose representation, particularly addressing OOV terms and semantic nuances, a \textbf{three-level matching strategy}, referred to as Cascading Vocabulary Resolution (CVR), is employed. The \textbf{first level} is \textbf{direct lookup}: it checks if the input token $t$ or its core concept has a directly corresponding CSL symbol $s_k$ within the \texttt{CNText2Sign} vocabulary. If such an entry exists, this symbol and its associated pose $P_k$ are directly selected. If direct lookup fails, the \textbf{second level}, \textbf{basic semantic retrieval}, is activated: the embedding $e(q)$ of the input token $q$ (typically $t$ itself) is used to search the \texttt{CNText2Sign} knowledge base $KB = \{(s_k, P_k)\}$ for the semantically closest CSL symbol $s^*$. The initial match $(s^*, P^*)$ is determined by maximizing semantic similarity, calculated using cosine similarity as defined in Equation~(\ref{eq:sim}).
\begin{equation}
    similarity(q, s_k) = \frac{e(q) \cdot e(s_k)}{||e(q)|| \, ||e(s_k)||},
    \label{eq:sim}
\end{equation}
where $e(q)$ is the embedding of the input token $q$, $e(s_k)$ is the embedding of a candidate CSL symbol $s_k$, $|| \cdot ||$ denotes the standard L2 norm, and $\cdot$ represents the dot product operation.
The selection is based on the criterion in Equation~(\ref{eq:argmax}).
\begin{equation}
    (s^*, P^*) = \underset{(s_k, P_k) \in KB}{\mathrm{argmax}} \; similarity(q, s_k),
    \label{eq:argmax}
\end{equation}
where $(s^*, P^*)$ represents the selected symbol-pose pair maximizing the similarity defined in Equation~(\ref{eq:sim}), and the maximization is performed over all pairs $(s_k, P_k)$ in the knowledge base $KB$.

In instances where neither direct lookup nor basic semantic retrieval produces a sufficiently high-confidence or contextually appropriate match, the \textbf{third level}, \textbf{LLM-driven synonym optimization with vocabulary integration}, is initiated. In this stage, the fine-tuned LLM actively generates or retrieves a diverse set of CSL synonyms or better alternative expressions for the input $q$ or the level-2 result $s^*$, referencing the $KB$ to ensure vocabulary validity and retrieve poses. The system then selects the most contextually appropriate synonym and its pose. The module ultimately generates the structured symbol-pose sequence $S_{pose} = \{(s_1, P_1), ..., (s_N, P_N)\}$ for the subsequent HSVSM module.

\vspace{-1em}
\subsection*{Hierarchical Sign Video Synthesis Module} 

The Hierarchical Sign Video Synthesis Module renders the input symbol-pose sequence $S_{pose} = \{(s_1, P_1), ..., (s_N, P_N)\}$, where each $P_i$ represents a \textbf{2D skeletal pose sequence} associated with symbol $s_i$, into a realistic \textbf{sign language performance}. It operates hierarchically to achieve both kinematic accuracy and visual fidelity. Initially, the \textbf{pose-conditioned generation} stage utilizes the input 2D pose sequences $P_i$ and employs methods based on ControlNet~\cite{Zhang2023ControlNet} as strong spatial conditioning to synthesize a base 2D visual representation that accurately matches the specified sign kinematics. Following this, the \textbf{detail refinement} stage applies Gen-3 Alpha~\cite{RunwayGen3Alpha2024} 
to enhance the realism and motion naturalness of this 2D representation, focusing on crucial facial expressions and intricate hand shapes, while ensuring temporally smooth transitions. The module ultimately outputs a coherent, high-fidelity sign language performance. This performance can be directly realized as a standard 2D video stream. Alternatively, for 3D applications, the \textbf{2D video frames} generated by the module serve as input to the Unique3D method~\cite{Li2023Unique3D} to produce a high-quality 3D virtual avatar, specifically a \textbf{textured mesh}, from the 2D input, allowing expressive sign language performance of 3D avatars in immersive environments.

\begin{figure*}[h] 
    \centering
    \includegraphics[width=\textwidth]{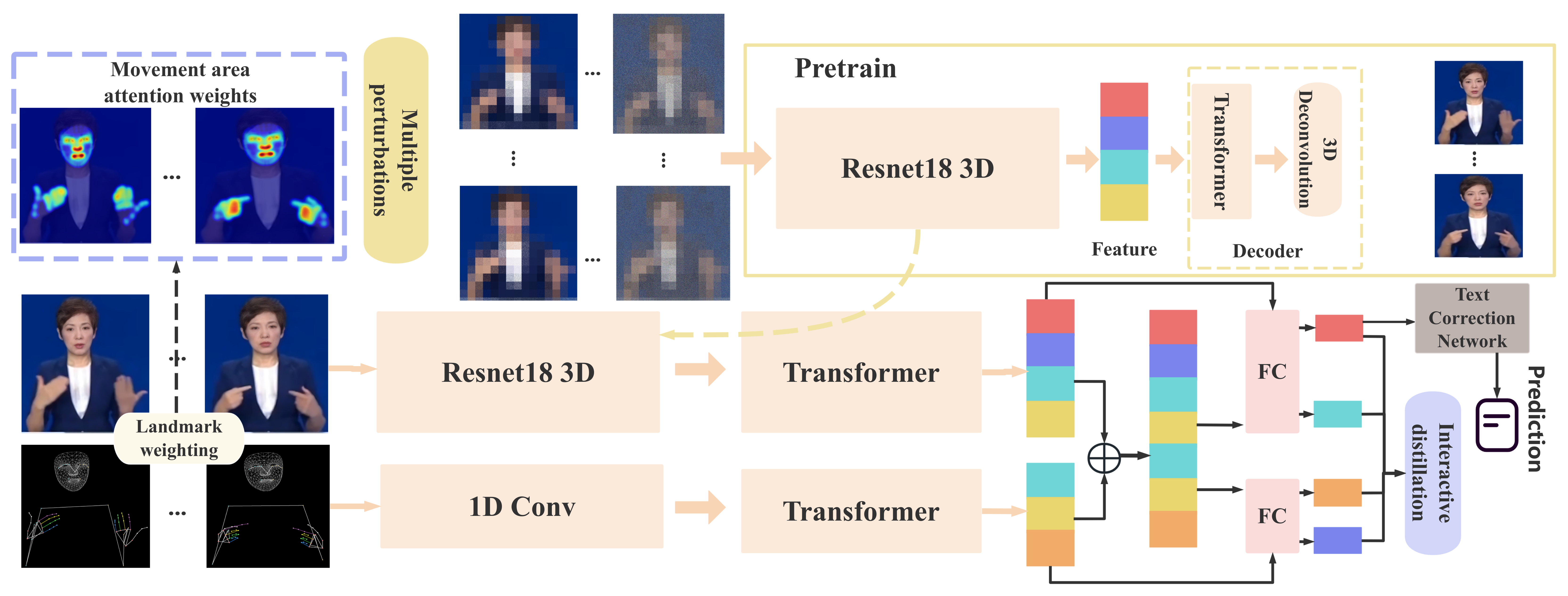}
    \vspace{-2em}
    \caption{An overview of SignMST-C, starts with video frames and landmarks processed through 3D ResNet18 and 1D convolution for 
    spatial-temporal and geometric features.}
     \label{fig:SLT}
  \end{figure*}

\vspace{-1em}
\subsection*{SLT Framework}
\noindent\hspace{1em}
We propose a SignMST-C (\textbf{Sign} \textbf{M}odel \textbf{S}elf-\textbf{T}ranslation with \textbf{C}orrection) 
framework based on multimodal fusion, as seen in Figure ~\ref{fig:SLT}, which combines self-supervised pretraining for rapid motion video semantic reconstruction, multimodal feature fusion of video and landmark data, end-to-end video-to-text translation, and a text correction network. The framework aims to improve the accuracy and timeliness of translating sign language videos into natural language text. Below are detailed descriptions and formula definitions for each model.

\noindent
\textbf{Self-Supervised Pretraining for Rapid Motion Video Semantic Reconstruction}

To enhance the model's attention to dynamic regions in sign language videos, we propose a self-supervised pretraining method. By introducing multiple perturbations and employing a landmark-based weighted perturbation strategy, the model is guided to prioritize rapidly moving regions during reconstruction. Specifically, we utilize a ResNet18-3D convolutional network for feature extraction and a lightweight decoder to reconstruct video content, emphasizing critical dynamic features. During input generation, perturbations applied to frames \( v_t \) include Pixel Shuffling to disrupt spatial structures, Random Pixel Replacement simulating blur effects, Block Occlusion introducing zero-value regions, Local Gaussian Noise mimicking capture interference, and Temporal Sequence Shuffling to disrupt the original temporal ordering.

In the process of constructing self-supervised input, we use landmark recognition to identify fast-moving 
regions in video frames, applying significantly higher perturbation weights to these critical areas. The detailed process is as follows: 

Identify fast-moving regions by calculating the movement speed between consecutive frames at landmark points.

For each landmark point \((x_i, y_i)\):
\begin{equation}
    S_i = \sqrt{(x_i - x_{i-1})^2 + (y_i - y_{i-1})^2},
\end{equation}
Where \( S_i > \theta \), this landmark point is defined as a fast-moving region, and the surrounding area with a radius of \( r \) is expanded, defined as:
\begin{align}
    M_t = \{ (x, y) \mid \sqrt{(x - x_i)^2 + (y - y_i)^2} \leq r, \notag \\
    \forall (x_i, y_i) \in L_t, S_i > \theta \},
\end{align}

For pixels in the fast-moving regions (including landmarks and surrounding areas), a large perturbation weight 
\( w_{\text{large}} \) is used. For other regions, a small perturbation weight \( w_{\text{small}} \) is applied. 
The weighted perturbed input is defined as:
\begin{equation}
    v_t' = w_{\text{large}} f_{\text{large}}(v_t, M_t) + w_{\text{small}} f_{\text{small}}(v_t, M_t'),
\end{equation}
where \( f_{\text{large}} \) and \( f_{\text{small}} \) represent large and small perturbation operations, \( || \cdot ||_2 \) denotes the L2 norm, and \( T \) represents the time step. Additionally, \( w_{\text{large}} + w_{\text{small}} = 1 \).

We define the reconstruction loss function \( L_{\text{recon}} \) as the average pixel-wise error between the original video sequence and the reconstructed video sequence, ensuring that the reconstructed output retains the visual content of the original video as closely as possible:
\begin{equation}
L_{\text{recon}} = \frac{1}{T} \sum_{t=1}^{T} \left\| v_t - \hat{v}_t \right\|_2^2,
\end{equation}
where \( v_t \) is the pixel value matrix of the original video frame, \( \hat{v}_t \) is the pixel value matrix of the reconstructed video frame generated by the model.

\begin{table*}[!t]
    \centering
    \begin{tabular}{l c c c c c}
    \hline
    \multicolumn{6}{c}{\textbf{CNText2Sign Sign Language Production Results (Dev/Test)}\bigstrut} \\ \hline
    \textbf{Model} & \textbf{BLEU-1} & \textbf{BLEU-2} & \textbf{BLEU-3} & \textbf{BLEU-4} & \textbf{CER (\%)} \\
    Qwen2.5-3b-instruct-AuralLLM & 58.32/57.13 & 50.6/49.91 & 41.68/40.32 & 35.17/34.26 & 46.61/45.23 \\ 
    Qwen2.5-32b-instruct-AuralLLM & 66.16/64.94 & 58.7/56.52 & 50.2/49.22 & 43.5/42.34 & 40.5/38.92 \\ 
    GLM-4-9b-AuralLLM & 62.58/61.37 & 54.15/53.14 & 45.13/44.59 & 38.16/36.38 & 46.2/45.14 \\ 
    \textbf{GLM-4-AuralLLM} & \textbf{69.54/68.31} & \textbf{62.03/61.24} & \textbf{53.31/51.25} & \textbf{50.41/49.23} & \textbf{34.95/33.86} \\ \hline
    \multicolumn{6}{c}{\textbf{CNSign SLT Results (Dev/Test)}\bigstrut} \\ \hline
    \textbf{Model} & \textbf{BLEU-1} & \textbf{BLEU-2} & \textbf{BLEU-3} & \textbf{BLEU-4} & \textbf{ROUGE} \\
    SignMST & 35.47/33.98 & 23.13/22.21 & 17.42/16.23 & 13.94/11.45 & 34.80/32.86 \\ 
    \textbf{SignMST-C} & \textbf{44.37/42.73} & \textbf{33.21/31.84} & \textbf{26.81/24.63} & \textbf{22.43/20.74} & \textbf{44.98/42.36} \\ \hline
    \end{tabular}
    \caption{Benchmark results for AuraLLM (SLP on CNText2Sign) and SignMST-C vs. SignMST (SLT on CNSign). Best  in bold.}
    \vspace{-2.5em}
    \label{table2}
\end{table*}

\noindent
\textbf{Multimodal Module for SLT}

In the multimodal module of the SLT framework, video frames are processed by a ResNet18 3D network to generate Conv\_Feature, capturing spatiotemporal dynamics, while landmark data is transformed via a 1D convolutional layer into 1D Conv Feature, encoding geometric positional variations. These features are concatenated and fused using a Transformer's multi-layer self-attention mechanism to obtain the multimodal representation Tran\_Feature. To ensure temporal consistency and semantic alignment, three distillation losses and a cross-entropy loss are designed, guiding the model towards robust spatiotemporal and structural comprehension.





In the process of multimodal fusion, the video and landmark features are derived from different networks. To maintain temporal consistency between these features, we introduce three distillation losses to ensure that the temporal structures of the features are aligned before and after fusion, thus avoiding information distortion or conflict. These losses measure the similarity between feature distributions using Kullback-Leibler (KL) Divergence:
\vspace{-0.5em}
\begin{equation}
    L_{\text{self-KL}} = \frac{1}{T} \sum_{t=1}^{T} \text{KL}\left(P_{3D}(t) \parallel P_{\text{T}}(t)\right),
    \vspace{-0.5em}
\end{equation}    
where \( P_{3D}(t) \) is the feature distribution of the 3D convolution, and \( P_{\text{T}}(t) \) is the fused feature distribution from the Transformer
    
\( L_{\text{LM-T-KL}} \) loss ensures that the fused features of landmarks and video remain consistent across time steps, enabling the model to capture dynamic information in sign language videos effectively:
\begin{align}
\vspace{-0.5em}
    L_{\text{LM-T-KL}} = \frac{1}{T} \sum_{t=1}^{T} \text{KL}(P_{\text{LT}}(t) \parallel P_{\text{VT}}(t)),
    \vspace{-0.5em}
\end{align}    
where \( P_{\text{LT}}(t) \)represents the feature distribution obtained from the landmark transformer and  \( P_{\text{VT}}(t) \) represents the corresponding feature distribution from the video transformer. 


The final total loss function \( L_{\text{total}} \) for the multimodal module combines the three distillation losses with cross-entropy loss to optimize feature alignment and semantic representation:
\begin{equation}
    L_{\text{total}} = L_{\text{self-KL}} + L_{\text{LM-Video-KL}} + L_{\text{LM-T-KL}} + L_{\text{CE}}.
    \end{equation}
    

\noindent
\textbf{Text Correction Network}

A Text Correction Network module is incorporated for the post-processing of preliminary translations generated by the Sign Language Translation model. Its objective is to mitigate translation errors, including but not limited to inadequate word order, omissions, redundancies, and substitutions. The training paradigm for this network leverages a self-supervised methodology adapted from KD-MSLRT ~\cite{li2025kdmslrtlightweightsignlanguage}. This entails the generation of paired training data wherein ground-truth reference sentences are subjected to synthetically induced perturbations. These perturbations encompass random word shuffling, deletion, substitution, and insertion operations, designed to emulate characteristic error distributions inherent in machine-generated translations. The network is subsequently optimized to learn the mapping function from these corrupted sequences back to their original, error-free forms. This process endows the network with the enhanced capacity to rectify translation inaccuracies, thereby improving the grammatical coherence and semantic fidelity of the final output text. 



\section{Experiment}
\label{fif:experiment}
\subsection*{Experiment Setup}
Experiments were conducted on NVIDIA A100-80G GPUs. We utilized our proposed CNText2Sign and CNSign datasets with their predefined 80\%/10\%/10\% splits, alongside the PHOENIX2014-T benchmark for SLT comparison. For SLP, AuraLLM variants employed different LLM backbones (e.g., GLM-4, Qwen2.5) fine-tuned via LoRA. For SLT, SignMST-C integrates ResNet18-3D, 1D convolutions, Transformer fusion, and a Text Correction Network. The SignMST-C model was trained using the AdamW optimizer with an initial learning rate of 0.0001, batch size of 16, and employed a linear learning rate decay schedule.

\textbf{Metric} In this study, we utilize BLEU (Bilingual Evaluation Understudy), CER (Character Error Rate), and ROUGE (Recall-Oriented Understudy for Gisting Evaluation) as the primary and standard evaluation metrics for the SLP and SLT tasks. 

\begin{figure}[t!] 
  \centering
  \includegraphics[width=0.5\textwidth]{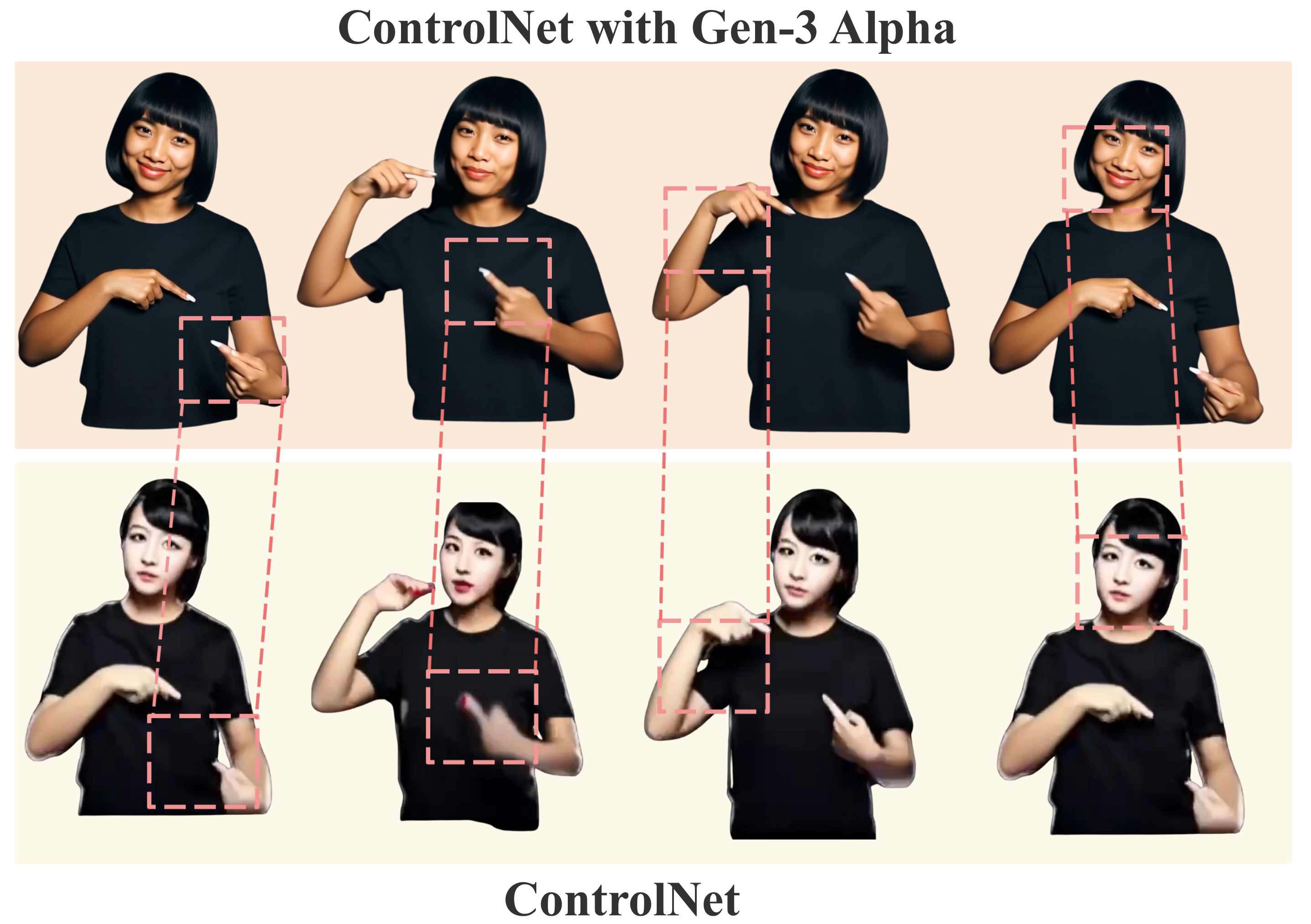}
  \vspace{-2em}
  \caption{Demonstrating the enhanced detail and realism from Gen-3 Alpha refinement (top) over baseline ControlNet generation (bottom) for sign language video.}
  \label{Gen-3}
  \vspace{-2.1em}
\end{figure}

\subsection*{Result}

Table ~\ref{table2} summarizes the benchmark performance of our proposed AuraLLM and SignMST-C models on the CNText2Sign and CNSign datasets, respectively. For SLP, the AuraLLM variant using the GLM-4 backbone achieves the best results (BLEU-4 50.41/49.23 on Dev/Test), evaluated via direct natural language to CSL symbol mapping accuracy (BLEU/CER) enabled by CNText2Sign, thus avoiding back-translation limitations. For SLT on CNSign, SignMST-C significantly surpasses a baseline lacking the text correction network, reaching a BLEU-4 of 22.43/20.74 and ROUGE of 44.98/42.36 (Dev/Test), demonstrating the effectiveness of the integrated correction module.Furthermore, when evaluated on the public Phoenix2014-T benchmark against existing methods as shown is Table \ref{tab:comparison}, SignMST-C establishes new state-of-the-art performance for the sign-to-text task, achieving a BLEU-4 score of 32.08.

\subsection*{Ablation Studies}

The visual comparison in Figure \ref{Gen-3} clearly shows that adding the Gen-3 Alpha refinement stage (top row) on top of ControlNet (bottom row) significantly improves the quality of the generated sign language video. Key improvements are evident in the naturalness of facial expressions and the clarity of hand gestures, making the final output frames more realistic and expressive

\begin{table}[t!]
    \centering
    \small
    \setlength{\tabcolsep}{3.5pt}
    \scalebox{0.9}{
    \begin{tabular}{l c c c c c}
        \hline
        \textbf{Model} & \textbf{ROUGE} & \textbf{BLEU-1} & \textbf{BLEU-2} & \textbf{BLEU-3} & \textbf{BLEU-4} \\
        \hline
        \multicolumn{6}{c}{\textbf{Gloss-based}} \\
        \hline
        SL-Transformer~\cite{Camgoz2020SignLanguageTransformers} & - & 46.61 & 33.73 & 26.19 & 21.32 \\
        BN-TIN-TransI+BT~\cite{Zhou2021ImprovingSignLanguageTranslation} & 49.54 & 50.80 & 37.75 & 29.72 & 24.32 \\
        MMTLB~\cite{chen2023simplemultimodalitytransferlearning} & 52.65 & 53.97 & 41.75 & 33.84 & 28.39 \\
        SLTU$_{SEQ}$~\cite{zhang2023sltunetsimpleunifiedmodel} & 52.11 & 52.92 & 41.76 & 33.99 & 28.47 \\
        TwoStream-SLT~\cite{chen2023twostreamnetworksignlanguage} & 53.48 & 54.90 & 42.43 & 34.46 & 28.95 \\
        \hline
        \multicolumn{6}{c}{\textbf{Gloss-free}} \\
        \hline
        NSLT~\cite{camgoz2020signlanguagetransformersjoint} & 30.70 & 29.86 & 17.52 & 11.96 & 9.00 \\
        TSPNet~\cite{YU2020104652} & 34.96 & 36.10 & 23.12 & 16.88 & 13.41 \\
        GASLT~\cite{yin2023glossattentionglossfreesign} & 39.86 & 39.07 & 26.74 & 21.86 & 15.74 \\
        GFSLT~\cite{zhou2023glossfreesignlanguagetranslation} & 40.93 & 41.39 & 31.00 & 24.20 & 19.66 \\
        GFSLT-VLP~\cite{zhou2023glossfreesignlanguagetranslation} & 42.49 & 43.71 & 33.18 & 26.11 & 21.44 \\
        SignLLM~\cite{gong2024llmsgoodsignlanguage} & 44.49 & 45.21 & 34.78 & 28.05 & 23.49 \\
        Fia-LLM~\cite{chen-etal-2024-factorized} & 45.27 & 46.29 & 35.33 & 28.03 & 23.09 \\
        Sign2GPT-FGP~\cite{wong2024sign2gptleveraginglargelanguage} & 48.90 & 49.54 & 35.96 & 28.83 & 22.52 \\
        GFSLT-VLP SignCL~\cite{ye2024improvingglossfreesignlanguage} & 49.04 & 49.76 & 36.85 & 29.97 & 22.74 \\
        LLaVA-SLT \cite{liang2024llavasltvisuallanguagetuning} & 50.44 & 51.20 & 37.51 & 29.39 & 23.43 \\
        \hline
        \textbf{SignMST-C (Ours)} & \textbf{56.13} & \textbf{57.63} & \textbf{45.32} & \textbf{37.21} & \textbf{32.08} \\
        \hline
    \end{tabular}}
    \caption{Comparison of State-of-the-Art Methods on the Phoenix2014-T Dataset (Test dataset) for the Sign-to-Text Task. Results better than the SOTAs are in bold.}
    \vspace{-2.5em}
    \label{tab:comparison}
\end{table}

Table \ref{table4} reveals a stark performance gap on the NL-to-CSL symbol mapping task. Directly applied base LLMs demonstrate poor performance; for instance, GLM-4-0520 achieves a BLEU-4 score of only 15.64. In contrast, our proposed GLM-4-AuralLLM reaches a significantly higher BLEU-4 score of 34.95, according to this table. This large difference underscores the inadequacy of direct application and validates that AuraLLM's core innovations are essential to effectively address this challenge: specifically, its decoupled semantic-to-representation approach and its specialized multi-level mapping strategy that integrates CVR and retrieval mechanisms.
\begin{table}[H]
    \centering
    \scalebox{0.9}{
    \begin{tabular}{l c c c c}
        \hline
        \textbf{Model} & \textbf{BLEU-1} & \textbf{BLEU-2} & \textbf{BLEU-3} & \textbf{BLEU-4} \\ \hline
        GPT-4o & 32.47 & 28.79 & 24.43 & 19.87 \\ 
        GLM-4 & 29.13 & 24.59 & 20.11 & 15.64 \\ 
        Qwen2.5-32-instruct & 23.51 & 19.91 & 15.83 & 11.27 \\ 
        \textbf{GLM-4-AuralLLM} & \textbf{69.54} & \textbf{62.03} & \textbf{53.31} & \textbf{34.95} \\ \hline
    \end{tabular}
    }
    \caption{Comparison of base LLMs vs. Glm-4-0520-AuralLLM performance. Best results in bold.}
    \label{table4}
    \vspace{-2em}
\end{table}

Ablation results in Table \ref{tab:ablation_slp_components} (CER, \%) demonstrate the positive contribution of Cascading Vocabulary Resolution (CVR), LoRA fine-tuning, and CVR's Embedding Search component. The optimal CER of 34.95\% is achieved only when all three components are combined, as removing any single component significantly degrades performance. This validates the synergistic effectiveness of integrating these elements for the NL-to-CSL task.

\begin{table}[h]
  \centering
  \begin{tabular}{c c c c}
    \hline
    \textbf{CVR} & \textbf{LoRA} & \textbf{Embedding Search} & \textbf{CER (\%)} \\ 
    \hline
    \checkmark & \texttimes & \texttimes & 41.83 \\
    \checkmark & \checkmark & \texttimes & 45.83 \\ 
    \checkmark & \texttimes & \checkmark & 38.64 \\
    \texttimes & \checkmark & \texttimes & 56.73 \\
    \checkmark & \checkmark & \checkmark & \textbf{34.95} \\
    \hline
  \end{tabular}
  \caption{Effect of AuraLLM's Cascading Vocabulary Resolution (CVR), its core Embedding Search component, and LoRA fine-tuning on Character Error Rate (CER, \%). CVR refers to the multi-level strategy for NL-to-CSL symbol mapping, while Embedding Search is the semantic retrieval part within CVR. Best result in bold.}
  \label{tab:ablation_slp_components} 
  \vspace{-2.5em} 
\end{table}

The ablation study for SignMST-C components on the CNSign test set as shown in Table \ref{table6}, ROUGE scores, shows Self-supervised Pretraining provides a strong foundation, significantly boosting the ROUGE score; adding it elevates performance from a ROUGE of 32.54 to 42.36 when combined with the other two components. The Text Correction Network proves crucial for refining the final output, yielding substantial gains especially over robust features, as demonstrated by the score increasing from 32.86 to the optimal 42.36 upon its inclusion. Multimodal Fusion also contributes positively—its addition improves the ROUGE score from 37.53 to 42.36 when integrated with pretraining and correction. The study highlights that the synergy of all three components achieves the best performance (42.36), far surpassing partial configurations, emphasizing the value of combining robust feature learning with explicit output correction for high-quality SLT.

\begin{table}[h]
    \centering
    \scalebox{0.9}{
    \begin{tabular}{c c c c}
        \hline
        \textbf{Pretrain} & \textbf{Landmark} & \textbf{Text Correction Network} & \textbf{ROUGE} \\ \hline
        \checkmark & \texttimes & \texttimes & 30.13 \\ 
        \checkmark & \checkmark & \texttimes & 32.86 \\ 
        \checkmark & \texttimes & \checkmark & 37.53 \\ 
        \texttimes & \checkmark & \texttimes & 27.41 \\ 
        \texttimes & \texttimes & \checkmark & 31.13 \\ 
        \texttimes & \checkmark & \checkmark & 32.54 \\ 
        \checkmark & \checkmark & \checkmark & \textbf{42.36} \\ \hline
    \end{tabular}
    }
    \caption{Ablation study on SignMST-C components. Impact on ROUGE score from enabling Self-supervised Pretraining (Pretrain), Multimodal Fusion (Landmark), and the Text Correction Network. Best result in bold.}
    \label{table6}
    \vspace{-2.5em}
\end{table}

\section{Conclusion}
\label{sec:conclusion}

This research addresses key challenges in bidirectional Chinese Sign Language communication by contributing the CNText2Sign and CNSign datasets, providing a unified data foundation for CSL. Notably, CNText2Sign's vocabulary-to-pose mapping enables direct evaluation of SLP accuracy, overcoming the limitations of traditional back-translation. Building on this, the AuraLLM model achieves high-quality, controllable sign language production, while the SignMST-C model attains state-of-the-art performance in sign language translation, collectively advancing accessible communication technology.

\bibliographystyle{ACM-Reference-Format}
\bibliography{sample-base}


\begin{thebibliography}{59}


\ifx \showCODEN    \undefined \def \showCODEN     #1{\unskip}     \fi
\ifx \showISBNx    \undefined \def \showISBNx     #1{\unskip}     \fi
\ifx \showISBNxiii \undefined \def \showISBNxiii  #1{\unskip}     \fi
\ifx \showISSN     \undefined \def \showISSN      #1{\unskip}     \fi
\ifx \showLCCN     \undefined \def \showLCCN      #1{\unskip}     \fi
\ifx \shownote     \undefined \def \shownote      #1{#1}          \fi
\ifx \showarticletitle \undefined \def \showarticletitle #1{#1}   \fi
\ifx \showURL      \undefined \def \showURL       {\relax}        \fi
\providecommand\bibfield[2]{#2}
\providecommand\bibinfo[2]{#2}
\providecommand\natexlab[1]{#1}
\providecommand\showeprint[2][]{arXiv:#2}

\bibitem[Agris and Kraiss(2010)]%
        {vonagris2010signum}
\bibfield{author}{\bibinfo{person}{Uwe~Von Agris} {and} \bibinfo{person}{Klaus~F. Kraiss}.} \bibinfo{year}{2010}\natexlab{}.
\newblock \showarticletitle{Signum database: Video corpus for signer-independent continuous sign language recognition}. In \bibinfo{booktitle}{\emph{Workshop on Representation and Processing of Sign Languages}}. \bibinfo{pages}{243--246}.
\newblock


\bibitem[Bohacek and Hrúz(2022)]%
        {Bohacek2022SignPoseTransformer}
\bibfield{author}{\bibinfo{person}{M. Bohacek} {and} \bibinfo{person}{M. Hrúz}.} \bibinfo{year}{2022}\natexlab{}.
\newblock \showarticletitle{Sign Pose-Based Transformer for Word-Level Sign Language Recognition}. In \bibinfo{booktitle}{\emph{Proceedings of the IEEE/CVF Winter Conference on Applications of Computer Vision (WACV) Workshops}}. \bibinfo{pages}{182--191}.
\newblock
\href{https://doi.org/10.1109/WACVW53916.2022.00039}{doi:\nolinkurl{10.1109/WACVW53916.2022.00039}}


\bibitem[Bragg et~al\mbox{.}(2019)]%
        {Bragg2019SignLanguageRecognition}
\bibfield{author}{\bibinfo{person}{D. Bragg}, \bibinfo{person}{O. Koller}, \bibinfo{person}{M. Bellard}, \bibinfo{person}{L. Berke}, \bibinfo{person}{M. Saenz}, {and} \bibinfo{person}{R. Kushalnagar}.} \bibinfo{year}{2019}\natexlab{}.
\newblock \showarticletitle{Sign Language Recognition, Generation, and Translation: An Interdisciplinary Perspective}. In \bibinfo{booktitle}{\emph{Proceedings of the 21st International ACM SIGACCESS Conference on Computers and Accessibility (ASSETS)}}.
\newblock
\urldef\tempurl%
\url{https://dl.acm.org/doi/10.1145/3308561.3353802}
\showURL{%
\tempurl}


\bibitem[Brown et~al\mbox{.}(2020)]%
        {Brown2020LanguageModels}
\bibfield{author}{\bibinfo{person}{T.B. Brown}, \bibinfo{person}{B. Mann}, \bibinfo{person}{N. Ryder}, \bibinfo{person}{M. Subbiah}, \bibinfo{person}{J. Kaplan}, \bibinfo{person}{P. Dhariwal}, \bibinfo{person}{A. Neelakantan}, \bibinfo{person}{P. Shyam}, \bibinfo{person}{G. Sastry}, \bibinfo{person}{A. Askell}, \bibinfo{person}{S. Agarwal}, \bibinfo{person}{A. Herbert-Voss}, \bibinfo{person}{G. Krueger}, \bibinfo{person}{T. Henighan}, \bibinfo{person}{R. Child}, \bibinfo{person}{A. Ramesh}, \bibinfo{person}{D.M. Ziegler}, \bibinfo{person}{J. Wu}, \bibinfo{person}{C. Winter}, \bibinfo{person}{C. Hesse}, \bibinfo{person}{M. Chen}, \bibinfo{person}{E. Sigler}, \bibinfo{person}{M. Litwin}, \bibinfo{person}{S. Gray}, \bibinfo{person}{B. Chess}, \bibinfo{person}{J. Clark}, \bibinfo{person}{C. Berner}, \bibinfo{person}{S. McCandlish}, \bibinfo{person}{A. Radford}, \bibinfo{person}{I. Sutskever}, {and} \bibinfo{person}{D. Amodei}.} \bibinfo{year}{2020}\natexlab{}.
\newblock \showarticletitle{Language Models are Few-Shot Learners}. In \bibinfo{booktitle}{\emph{Advances in Neural Information Processing Systems 33: Annual Conference on Neural Information Processing Systems 2020 (NeurIPS 2020)}}, \bibfield{editor}{\bibinfo{person}{H.~Larochelle}, \bibinfo{person}{M.~Ranzato}, \bibinfo{person}{R.~Hadsell}, \bibinfo{person}{M.~Balcan}, {and} \bibinfo{person}{H.~Lin}} (Eds.). \bibinfo{publisher}{Curran Associates, Inc.}, \bibinfo{pages}{1877--1901}.
\newblock
\urldef\tempurl%
\url{https://proceedings.neurips.cc/paper/2020/file/1457c0d8e8c8a6d9b72dcd67a23a036a-Paper.pdf}
\showURL{%
\tempurl}


\bibitem[Camgoz et~al\mbox{.}(2020a)]%
        {camgoz2020signlanguagetransformersjoint}
\bibfield{author}{\bibinfo{person}{Necati~Cihan Camgoz}, \bibinfo{person}{Oscar Koller}, \bibinfo{person}{Simon Hadfield}, {and} \bibinfo{person}{Richard Bowden}.} \bibinfo{year}{2020}\natexlab{a}.
\newblock \bibinfo{title}{Sign Language Transformers: Joint End-to-end Sign Language Recognition and Translation}.
\newblock
\showeprint[arxiv]{2003.13830}~[cs.CV]
\urldef\tempurl%
\url{https://arxiv.org/abs/2003.13830}
\showURL{%
\tempurl}


\bibitem[Camgoz et~al\mbox{.}(2020b)]%
        {Camgoz2020SignLanguageTransformers}
\bibfield{author}{\bibinfo{person}{Necati~Cihan Camgoz}, \bibinfo{person}{Oscar Koller}, \bibinfo{person}{Simon Hadfield}, {and} \bibinfo{person}{Richard Bowden}.} \bibinfo{year}{2020}\natexlab{b}.
\newblock \showarticletitle{Sign Language Transformers: Joint End-to-End Sign Language Recognition and Translation}. In \bibinfo{booktitle}{\emph{Proceedings of the IEEE/CVF Conference on Computer Vision and Pattern Recognition (CVPR)}}. \bibinfo{pages}{10023--10033}.
\newblock
\href{https://doi.org/10.1109/CVPR42600.2020.01002}{doi:\nolinkurl{10.1109/CVPR42600.2020.01002}}


\bibitem[Cao et~al\mbox{.}(2017)]%
        {Cao2017OpenPose}
\bibfield{author}{\bibinfo{person}{Z. Cao}, \bibinfo{person}{G. Hidalgo}, \bibinfo{person}{T. Simon}, \bibinfo{person}{S.E. Wei}, {and} \bibinfo{person}{Y. Sheikh}.} \bibinfo{year}{2017}\natexlab{}.
\newblock \showarticletitle{OpenPose: Realtime Multi-Person 2D Pose Estimation Using Part Affinity Fields}. In \bibinfo{booktitle}{\emph{Proceedings of the IEEE Conference on Computer Vision and Pattern Recognition (CVPR)}}.
\newblock
\urldef\tempurl%
\url{https://openaccess.thecvf.com/content_CVPR_2017/html/Cao_OpenPose_Realtime_CVPR_2017_paper.html}
\showURL{%
\tempurl}


\bibitem[Chen et~al\mbox{.}(2022a)]%
        {Chen2022SimpleMultiModalityTransferLearning}
\bibfield{author}{\bibinfo{person}{Yutong Chen}, \bibinfo{person}{Fangyun Wei}, \bibinfo{person}{Xiao Sun}, \bibinfo{person}{Zhirong Wu}, {and} \bibinfo{person}{Stephen Lin}.} \bibinfo{year}{2022}\natexlab{a}.
\newblock \showarticletitle{A Simple Multi-Modality Transfer Learning Baseline for Sign Language Translation}. In \bibinfo{booktitle}{\emph{Proceedings of the IEEE/CVF Conference on Computer Vision and Pattern Recognition (CVPR)}}. \bibinfo{pages}{5120--5130}.
\newblock
\href{https://doi.org/10.1109/CVPR52688.2022.00505}{doi:\nolinkurl{10.1109/CVPR52688.2022.00505}}


\bibitem[Chen et~al\mbox{.}(2023a)]%
        {chen2023simplemultimodalitytransferlearning}
\bibfield{author}{\bibinfo{person}{Yutong Chen}, \bibinfo{person}{Fangyun Wei}, \bibinfo{person}{Xiao Sun}, \bibinfo{person}{Zhirong Wu}, {and} \bibinfo{person}{Stephen Lin}.} \bibinfo{year}{2023}\natexlab{a}.
\newblock \bibinfo{title}{A Simple Multi-Modality Transfer Learning Baseline for Sign Language Translation}.
\newblock
\showeprint[arxiv]{2203.04287}~[cs.CV]
\urldef\tempurl%
\url{https://arxiv.org/abs/2203.04287}
\showURL{%
\tempurl}


\bibitem[Chen et~al\mbox{.}(2022b)]%
        {Chen2022TwoStreamNetworkForSignLanguage}
\bibfield{author}{\bibinfo{person}{Yutong Chen}, \bibinfo{person}{Ronglai Zuo}, \bibinfo{person}{Fangyun Wei}, \bibinfo{person}{Yu Wu}, \bibinfo{person}{Shujie Liu}, {and} \bibinfo{person}{Brian Mak}.} \bibinfo{year}{2022}\natexlab{b}.
\newblock \showarticletitle{Two-Stream Network for Sign Language Recognition and Translation}. In \bibinfo{booktitle}{\emph{Advances in Neural Information Processing Systems (NeurIPS)}}, \bibfield{editor}{\bibinfo{person}{Alice~H. Oh}, \bibinfo{person}{Alekh Agarwal}, \bibinfo{person}{Danielle Belgrave}, {and} \bibinfo{person}{Kyunghyun Cho}} (Eds.).
\newblock
\urldef\tempurl%
\url{https://arxiv.org/abs/2211.12234}
\showURL{%
\tempurl}
\newblock
\shownote{Proceedings of NeurIPS 2022}.


\bibitem[Chen et~al\mbox{.}(2023b)]%
        {chen2023twostreamnetworksignlanguage}
\bibfield{author}{\bibinfo{person}{Yutong Chen}, \bibinfo{person}{Ronglai Zuo}, \bibinfo{person}{Fangyun Wei}, \bibinfo{person}{Yu Wu}, \bibinfo{person}{Shujie Liu}, {and} \bibinfo{person}{Brian Mak}.} \bibinfo{year}{2023}\natexlab{b}.
\newblock \bibinfo{title}{Two-Stream Network for Sign Language Recognition and Translation}.
\newblock
\showeprint[arxiv]{2211.01367}~[cs.CV]
\urldef\tempurl%
\url{https://arxiv.org/abs/2211.01367}
\showURL{%
\tempurl}


\bibitem[Chen et~al\mbox{.}(2024)]%
        {chen-etal-2024-factorized}
\bibfield{author}{\bibinfo{person}{Zhigang Chen}, \bibinfo{person}{Benjia Zhou}, \bibinfo{person}{Jun Li}, \bibinfo{person}{Jun Wan}, \bibinfo{person}{Zhen Lei}, \bibinfo{person}{Ning Jiang}, \bibinfo{person}{Quan Lu}, {and} \bibinfo{person}{Guoqing Zhao}.} \bibinfo{year}{2024}\natexlab{}.
\newblock \showarticletitle{Factorized Learning Assisted with Large Language Model for Gloss-free Sign Language Translation}. In \bibinfo{booktitle}{\emph{Proceedings of the 2024 Joint International Conference on Computational Linguistics, Language Resources and Evaluation (LREC-COLING 2024)}}. \bibinfo{publisher}{ELRA and ICCL}, \bibinfo{address}{Torino, Italia}, \bibinfo{pages}{7071--7081}.
\newblock
\urldef\tempurl%
\url{https://aclanthology.org/2024.lrec-main.620/}
\showURL{%
\tempurl}


\bibitem[Chowdhery et~al\mbox{.}(2022)]%
        {Chowdhery2022Palm}
\bibfield{author}{\bibinfo{person}{A. Chowdhery}, \bibinfo{person}{S. Narang}, \bibinfo{person}{J. Devlin}, \bibinfo{person}{M. Bosma}, \bibinfo{person}{G. Mishra}, \bibinfo{person}{A. Roberts}, \bibinfo{person}{P. Barham}, \bibinfo{person}{H.W. Chung}, \bibinfo{person}{C. Sutton}, \bibinfo{person}{S. Gehrmann}, \bibinfo{person}{P. Schuh}, \bibinfo{person}{K. Shi}, \bibinfo{person}{S. Tsvyashchenko}, \bibinfo{person}{J. Maynez}, \bibinfo{person}{A. Rao}, \bibinfo{person}{P. Barnes}, \bibinfo{person}{Y. Tay}, \bibinfo{person}{N. Shazeer}, \bibinfo{person}{V. Prabhakaran}, \bibinfo{person}{E. Reif}, \bibinfo{person}{N. Du}, \bibinfo{person}{B. Hutchinson}, \bibinfo{person}{R. Pope}, \bibinfo{person}{J. Bradbury}, \bibinfo{person}{J. Austin}, \bibinfo{person}{M. Isard}, \bibinfo{person}{G. Gur-Ari}, \bibinfo{person}{P. Yin}, \bibinfo{person}{T. Duke}, \bibinfo{person}{A. Levskaya}, \bibinfo{person}{S. Ghemawat}, \bibinfo{person}{S. Dev}, \bibinfo{person}{H. Michalewski}, \bibinfo{person}{X. Garcia},
  \bibinfo{person}{V. Misra}, \bibinfo{person}{K. Robinson}, \bibinfo{person}{L. Fedus}, \bibinfo{person}{D. Zhou}, \bibinfo{person}{D. Ippolito}, \bibinfo{person}{D. Luan}, \bibinfo{person}{H. Lim}, \bibinfo{person}{B. Zoph}, \bibinfo{person}{A. Spiridonov}, \bibinfo{person}{R. Sepassi}, \bibinfo{person}{D. Dohan}, \bibinfo{person}{S. Agrawal}, \bibinfo{person}{M. Omernick}, \bibinfo{person}{A.M. Dai}, \bibinfo{person}{T.S. Pillai}, \bibinfo{person}{M. Pellat}, \bibinfo{person}{A. Lewkowycz}, \bibinfo{person}{E. Moreira}, \bibinfo{person}{R. Child}, \bibinfo{person}{O. Polozov}, \bibinfo{person}{K. Lee}, \bibinfo{person}{Z. Zhou}, \bibinfo{person}{X. Wang}, \bibinfo{person}{B. Saeta}, \bibinfo{person}{M. Diaz}, \bibinfo{person}{O. Firat}, \bibinfo{person}{M. Catasta}, \bibinfo{person}{J. Wei}, \bibinfo{person}{K. Meier-Hellstern}, \bibinfo{person}{D. Eck}, \bibinfo{person}{J. Dean}, \bibinfo{person}{S. Petrov}, {and} \bibinfo{person}{N. Fiedel}.} \bibinfo{year}{2022}\natexlab{}.
\newblock \showarticletitle{PALM: Scaling Language Modeling with Pathways}.
\newblock \bibinfo{journal}{\emph{CoRR}}  \bibinfo{volume}{abs/2204.02311} (\bibinfo{year}{2022}).
\newblock
\urldef\tempurl%
\url{https://arxiv.org/abs/2204.02311}
\showURL{%
\tempurl}


\bibitem[Cihan~Camgoz et~al\mbox{.}(2018)]%
        {Camgoz2018NeuralSignLanguageTranslation}
\bibfield{author}{\bibinfo{person}{N. Cihan~Camgoz}, \bibinfo{person}{S. Hadfield}, \bibinfo{person}{O. Koller}, \bibinfo{person}{H. Ney}, {and} \bibinfo{person}{R. Bowden}.} \bibinfo{year}{2018}\natexlab{}.
\newblock \showarticletitle{Neural Sign Language Translation}. In \bibinfo{booktitle}{\emph{Proceedings of the IEEE Conference on Computer Vision and Pattern Recognition (CVPR)}}. \bibinfo{pages}{7784--7793}.
\newblock
\href{https://doi.org/10.1109/CVPR.2018.00795}{doi:\nolinkurl{10.1109/CVPR.2018.00795}}


\bibitem[Cox et~al\mbox{.}(2002)]%
        {Cox2002TESSA}
\bibfield{author}{\bibinfo{person}{S. Cox}, \bibinfo{person}{M. Lincoln}, \bibinfo{person}{J. Tryggvason}, \bibinfo{person}{M. Nakisa}, \bibinfo{person}{M. Wells}, \bibinfo{person}{M. Tutt}, {and} \bibinfo{person}{S. Abbott}.} \bibinfo{year}{2002}\natexlab{}.
\newblock \showarticletitle{TESSA, A System to Aid Communication with Deaf People}. In \bibinfo{booktitle}{\emph{Proceedings of the ACM International Conference on Assistive Technologies}}.
\newblock
\urldef\tempurl%
\url{https://dl.acm.org/doi/10.1145/571130.571160}
\showURL{%
\tempurl}


\bibitem[Duarte et~al\mbox{.}(2021a)]%
        {duarte2021how2signlargescalemultimodaldataset}
\bibfield{author}{\bibinfo{person}{Amanda Duarte}, \bibinfo{person}{Shruti Palaskar}, \bibinfo{person}{Lucas Ventura}, \bibinfo{person}{Deepti Ghadiyaram}, \bibinfo{person}{Kenneth DeHaan}, \bibinfo{person}{Florian Metze}, \bibinfo{person}{Jordi Torres}, {and} \bibinfo{person}{Xavier~Giro i Nieto}.} \bibinfo{year}{2021}\natexlab{a}.
\newblock \bibinfo{title}{How2Sign: A Large-scale Multimodal Dataset for Continuous American Sign Language}.
\newblock
\showeprint[arxiv]{2008.08143}~[cs.CV]
\urldef\tempurl%
\url{https://arxiv.org/abs/2008.08143}
\showURL{%
\tempurl}


\bibitem[Duarte et~al\mbox{.}(2021b)]%
        {Duarte2021How2Sign}
\bibfield{author}{\bibinfo{person}{A. Duarte}, \bibinfo{person}{S. Palaskar}, \bibinfo{person}{L. Ventura}, \bibinfo{person}{D. Ghadiyaram}, \bibinfo{person}{K. DeHaan}, \bibinfo{person}{F. Metze}, \bibinfo{person}{J. Torres}, {and} \bibinfo{person}{X.~Giro i Nieto}.} \bibinfo{year}{2021}\natexlab{b}.
\newblock \showarticletitle{How2Sign: A Large-Scale Multimodal Dataset for Continuous American Sign Language}. In \bibinfo{booktitle}{\emph{Proceedings of the IEEE/CVF Conference on Computer Vision and Pattern Recognition (CVPR)}}.
\newblock
\urldef\tempurl%
\url{https://openaccess.thecvf.com/content/CVPR2021/html/Duarte_How2Sign_A_Large-Scale_Multimodal_Dataset_for_Continuous_American_Sign_Language_CVPR_2021_paper.html}
\showURL{%
\tempurl}


\bibitem[Fang et~al\mbox{.}(2024)]%
        {fang2024signllmsignlanguagesproduction}
\bibfield{author}{\bibinfo{person}{Sen Fang}, \bibinfo{person}{Lei Wang}, \bibinfo{person}{Ce Zheng}, \bibinfo{person}{Yapeng Tian}, {and} \bibinfo{person}{Chen Chen}.} \bibinfo{year}{2024}\natexlab{}.
\newblock \bibinfo{title}{SignLLM: Sign Languages Production Large Language Models}.
\newblock
\showeprint[arxiv]{2405.10718}~[cs.CV]
\urldef\tempurl%
\url{https://arxiv.org/abs/2405.10718}
\showURL{%
\tempurl}


\bibitem[Gong et~al\mbox{.}(2024)]%
        {gong2024llmsgoodsignlanguage}
\bibfield{author}{\bibinfo{person}{Jia Gong}, \bibinfo{person}{Lin~Geng Foo}, \bibinfo{person}{Yixuan He}, \bibinfo{person}{Hossein Rahmani}, {and} \bibinfo{person}{Jun Liu}.} \bibinfo{year}{2024}\natexlab{}.
\newblock \bibinfo{title}{LLMs are Good Sign Language Translators}.
\newblock
\showeprint[arxiv]{2404.00925}~[cs.CV]
\urldef\tempurl%
\url{https://arxiv.org/abs/2404.00925}
\showURL{%
\tempurl}


\bibitem[Hanke et~al\mbox{.}(2020)]%
        {hanke2020extending}
\bibfield{author}{\bibinfo{person}{Tobias Hanke}, \bibinfo{person}{Michael Schulder}, \bibinfo{person}{Roland Konrad}, {and} \bibinfo{person}{Elias Jahn}.} \bibinfo{year}{2020}\natexlab{}.
\newblock \showarticletitle{Extending the Public DGS Corpus in Size and Depth}. In \bibinfo{booktitle}{\emph{LREC2020 - Workshop on the Representation and Processing of Sign Languages}}. \bibinfo{pages}{75--82}.
\newblock


\bibitem[Huang et~al\mbox{.}(2018)]%
        {huang2018video}
\bibfield{author}{\bibinfo{person}{J. Huang}, \bibinfo{person}{W. Zhou}, \bibinfo{person}{Q. Zhang}, \bibinfo{person}{H. Li}, {and} \bibinfo{person}{W. Li}.} \bibinfo{year}{2018}\natexlab{}.
\newblock \showarticletitle{Video-based Sign Language Recognition without Temporal Segmentation}. In \bibinfo{booktitle}{\emph{Proceedings of the AAAI Conference on Artificial Intelligence}}.
\newblock


\bibitem[Huang et~al\mbox{.}(2021)]%
        {Huang2021FastAndHighQualitySignLanguage}
\bibfield{author}{\bibinfo{person}{W. Huang}, \bibinfo{person}{W. Pan}, \bibinfo{person}{Z. Zhao}, {and} \bibinfo{person}{Q. Tian}.} \bibinfo{year}{2021}\natexlab{}.
\newblock \showarticletitle{Towards Fast and High-Quality Sign Language Production}. In \bibinfo{booktitle}{\emph{Proceedings of the 29th ACM International Conference on Multimedia (ACM MM)}}. \bibinfo{publisher}{ACM}.
\newblock
\urldef\tempurl%
\url{https://dl.acm.org/doi/10.1145/3474085.3475239}
\showURL{%
\tempurl}


\bibitem[Karpouzis et~al\mbox{.}(2007)]%
        {Karpouzis2007GreekSignLanguage}
\bibfield{author}{\bibinfo{person}{K. Karpouzis}, \bibinfo{person}{G. Caridakis}, \bibinfo{person}{S.E. Fotinea}, {and} \bibinfo{person}{E. Efthimiou}.} \bibinfo{year}{2007}\natexlab{}.
\newblock \showarticletitle{Educational Resources and Implementation of a Greek Sign Language Synthesis Architecture}.
\newblock \bibinfo{journal}{\emph{Computers \& Education}} \bibinfo{volume}{48}, \bibinfo{number}{1} (\bibinfo{year}{2007}), \bibinfo{pages}{35--52}.
\newblock
\href{https://doi.org/10.1016/j.compedu.2005.03.003}{doi:\nolinkurl{10.1016/j.compedu.2005.03.003}}


\bibitem[Ko et~al\mbox{.}(2019)]%
        {Ko2019NeuralSignLanguageTranslation}
\bibfield{author}{\bibinfo{person}{S.K. Ko}, \bibinfo{person}{C.J. Kim}, \bibinfo{person}{H. Jung}, {and} \bibinfo{person}{C. Cho}.} \bibinfo{year}{2019}\natexlab{}.
\newblock \showarticletitle{Neural Sign Language Translation based on Human Keypoint Estimation}.
\newblock \bibinfo{journal}{\emph{Applied Sciences}} \bibinfo{volume}{9}, \bibinfo{number}{20} (\bibinfo{year}{2019}), \bibinfo{pages}{4307}.
\newblock
\href{https://doi.org/10.3390/app9204307}{doi:\nolinkurl{10.3390/app9204307}}


\bibitem[Kreutzer et~al\mbox{.}(2019)]%
        {Kreutzer2019JoeyNMT}
\bibfield{author}{\bibinfo{person}{J. Kreutzer}, \bibinfo{person}{J. Bastings}, {and} \bibinfo{person}{S. Riezler}.} \bibinfo{year}{2019}\natexlab{}.
\newblock \showarticletitle{Joey NMT: A Minimalist NMT Toolkit for Novices}. In \bibinfo{booktitle}{\emph{Proceedings of the Conference on Empirical Methods in Natural Language Processing (EMNLP)}}.
\newblock
\urldef\tempurl%
\url{https://aclanthology.org/D19-5262}
\showURL{%
\tempurl}


\bibitem[Li et~al\mbox{.}(2020)]%
        {Li2020TSPNet}
\bibfield{author}{\bibinfo{person}{Dongxu Li}, \bibinfo{person}{Chenchen Xu}, \bibinfo{person}{Xin Yu}, \bibinfo{person}{Kaihao Zhang}, \bibinfo{person}{Benjamin Swift}, \bibinfo{person}{Hanna Suominen}, {and} \bibinfo{person}{Hongdong Li}.} \bibinfo{year}{2020}\natexlab{}.
\newblock \showarticletitle{TSPNet: Hierarchical Feature Learning via Temporal Semantic Pyramid for Sign Language Translation}. In \bibinfo{booktitle}{\emph{Advances in Neural Information Processing Systems (NeurIPS)}}, Vol.~\bibinfo{volume}{33}. \bibinfo{pages}{12034--12045}.
\newblock
\urldef\tempurl%
\url{https://arxiv.org/abs/2009.08687}
\showURL{%
\tempurl}


\bibitem[Li et~al\mbox{.}(2025)]%
        {li2025kdmslrtlightweightsignlanguage}
\bibfield{author}{\bibinfo{person}{Yulong Li}, \bibinfo{person}{Bolin Ren}, \bibinfo{person}{Ke Hu}, \bibinfo{person}{Changyuan Liu}, \bibinfo{person}{Zhengyong Jiang}, \bibinfo{person}{Kang Dang}, {and} \bibinfo{person}{Jionglong Su}.} \bibinfo{year}{2025}\natexlab{}.
\newblock \bibinfo{title}{KD-MSLRT: Lightweight Sign Language Recognition Model Based on Mediapipe and 3D to 1D Knowledge Distillation}.
\newblock
\showeprint[arxiv]{2501.02321}~[cs.CY]
\urldef\tempurl%
\url{https://arxiv.org/abs/2501.02321}
\showURL{%
\tempurl}


\bibitem[Liang et~al\mbox{.}(2024)]%
        {liang2024llavasltvisuallanguagetuning}
\bibfield{author}{\bibinfo{person}{Han Liang}, \bibinfo{person}{Chengyu Huang}, \bibinfo{person}{Yuecheng Xu}, \bibinfo{person}{Cheng Tang}, \bibinfo{person}{Weicai Ye}, \bibinfo{person}{Juze Zhang}, \bibinfo{person}{Xin Chen}, \bibinfo{person}{Jingyi Yu}, {and} \bibinfo{person}{Lan Xu}.} \bibinfo{year}{2024}\natexlab{}.
\newblock \bibinfo{title}{LLaVA-SLT: Visual Language Tuning for Sign Language Translation}.
\newblock
\showeprint[arxiv]{2412.16524}~[cs.CV]
\urldef\tempurl%
\url{https://arxiv.org/abs/2412.16524}
\showURL{%
\tempurl}


\bibitem[Lugaresi et~al\mbox{.}(2019)]%
        {lugaresi2019mediapipeframeworkbuildingperception}
\bibfield{author}{\bibinfo{person}{Camillo Lugaresi}, \bibinfo{person}{Jiuqiang Tang}, \bibinfo{person}{Hadon Nash}, \bibinfo{person}{Chris McClanahan}, \bibinfo{person}{Esha Uboweja}, \bibinfo{person}{Michael Hays}, \bibinfo{person}{Fan Zhang}, \bibinfo{person}{Chuo-Ling Chang}, \bibinfo{person}{Ming~Guang Yong}, \bibinfo{person}{Juhyun Lee}, \bibinfo{person}{Wan-Teh Chang}, \bibinfo{person}{Wei Hua}, \bibinfo{person}{Manfred Georg}, {and} \bibinfo{person}{Matthias Grundmann}.} \bibinfo{year}{2019}\natexlab{}.
\newblock \bibinfo{title}{MediaPipe: A Framework for Building Perception Pipelines}.
\newblock
\showeprint[arxiv]{1906.08172}~[cs.DC]
\urldef\tempurl%
\url{https://arxiv.org/abs/1906.08172}
\showURL{%
\tempurl}


\bibitem[Mazumder et~al\mbox{.}(2021)]%
        {Mazumder2021TranslatingSignLanguage}
\bibfield{author}{\bibinfo{person}{S. Mazumder}, \bibinfo{person}{R. Mukhopadhyay}, \bibinfo{person}{V.P. Namboodiri}, {and} \bibinfo{person}{C.V. Jawahar}.} \bibinfo{year}{2021}\natexlab{}.
\newblock \showarticletitle{Translating Sign Language Videos to Talking Faces}. In \bibinfo{booktitle}{\emph{Proceedings of the Twelfth Indian Conference on Computer Vision, Graphics and Image Processing (ICVGIP ’21)}}. \bibinfo{publisher}{Association for Computing Machinery}, \bibinfo{address}{New York, NY, USA}.
\newblock
\href{https://doi.org/10.1145/3490035.3490286}{doi:\nolinkurl{10.1145/3490035.3490286}}


\bibitem[McDonald et~al\mbox{.}(2016)]%
        {McDonald2016AutomatedTechnique}
\bibfield{author}{\bibinfo{person}{J. McDonald}, \bibinfo{person}{R. Wolfe}, \bibinfo{person}{J. Schnepp}, \bibinfo{person}{J. Hochgesang}, \bibinfo{person}{D.G. Jamrozik}, \bibinfo{person}{M. Stumbo}, \bibinfo{person}{L. Berke}, \bibinfo{person}{M. Bialek}, {and} \bibinfo{person}{F. Thomas}.} \bibinfo{year}{2016}\natexlab{}.
\newblock \showarticletitle{Automated Technique for Real-Time Production of Lifelike Animations of American Sign Language}.
\newblock \bibinfo{journal}{\emph{Universal Access in the Information Society (UAIS)}} \bibinfo{volume}{15}, \bibinfo{number}{4} (\bibinfo{year}{2016}), \bibinfo{pages}{755--769}.
\newblock
\href{https://doi.org/10.1007/s10209-016-0499-1}{doi:\nolinkurl{10.1007/s10209-016-0499-1}}


\bibitem[Nations(2023)]%
        {un_wfd_2023}
\bibfield{author}{\bibinfo{person}{United Nations}.} \bibinfo{year}{2023}\natexlab{}.
\newblock \bibinfo{title}{International Day of Sign Languages}.
\newblock
\urldef\tempurl%
\url{https://www.un.org/en/observances/sign-languages-day}
\showURL{%
\tempurl}
\newblock
\shownote{Accessed: 2024-11-14}.


\bibitem[Neidle and Vogler(2012)]%
        {neidle2012new}
\bibfield{author}{\bibinfo{person}{C. Neidle} {and} \bibinfo{person}{C. Vogler}.} \bibinfo{year}{2012}\natexlab{}.
\newblock \showarticletitle{A New Web Interface to Facilitate Access to Corpora: Development of the ASLLRP Data Access Interface (DAI)}. In \bibinfo{booktitle}{\emph{Proceedings of the 5th Workshop on the Representation and Processing of Sign Languages: Interactions between Corpus and Lexicon}}. \bibinfo{publisher}{LREC}.
\newblock


\bibitem[Papastratis et~al\mbox{.}(2021)]%
        {Papastratis}
\bibfield{author}{\bibinfo{person}{Ilias Papastratis}, \bibinfo{person}{Christos Chatzikonstantinou}, \bibinfo{person}{Dimitrios Konstantinidis}, \bibinfo{person}{Kosmas Dimitropoulos}, {and} \bibinfo{person}{Petros Daras}.} \bibinfo{year}{2021}\natexlab{}.
\newblock \showarticletitle{Artificial Intelligence Technologies for Sign Language}.
\newblock \bibinfo{journal}{\emph{Sensors}} \bibinfo{volume}{21}, \bibinfo{number}{17} (\bibinfo{year}{2021}).
\newblock
\showISSN{1424-8220}
\href{https://doi.org/10.3390/s21175843}{doi:\nolinkurl{10.3390/s21175843}}


\bibitem[Paszke et~al\mbox{.}(2017)]%
        {Paszke2017AutoDiffPyTorch}
\bibfield{author}{\bibinfo{person}{A. Paszke}, \bibinfo{person}{S. Gross}, \bibinfo{person}{S. Chintala}, \bibinfo{person}{G. Chanan}, \bibinfo{person}{E. Yang}, \bibinfo{person}{Z. DeVito}, \bibinfo{person}{Z. Lin}, \bibinfo{person}{A. Desmaison}, \bibinfo{person}{L. Antiga}, {and} \bibinfo{person}{A. Lerer}.} \bibinfo{year}{2017}\natexlab{}.
\newblock \showarticletitle{Automatic Differentiation in PyTorch}. In \bibinfo{booktitle}{\emph{NIPS Autodiff Workshop}}.
\newblock


\bibitem[Radford et~al\mbox{.}(2019)]%
        {Radford2019LanguageModels}
\bibfield{author}{\bibinfo{person}{A. Radford}, \bibinfo{person}{J. Wu}, \bibinfo{person}{R. Child}, \bibinfo{person}{D. Luan}, \bibinfo{person}{D. Amodei}, \bibinfo{person}{I. Sutskever}, {and} \bibinfo{person}{et al.}} \bibinfo{year}{2019}\natexlab{}.
\newblock \bibinfo{title}{Language models are unsupervised multitask learners}.
\newblock \bibinfo{howpublished}{OpenAI blog}.
\newblock
\urldef\tempurl%
\url{https://openai.com/blog/language-unsupervised}
\showURL{%
\tempurl}
\newblock
\shownote{p. 9}.


\bibitem[Raffel et~al\mbox{.}(2020)]%
        {Raffel2020ExploringTransferLearning}
\bibfield{author}{\bibinfo{person}{C. Raffel}, \bibinfo{person}{N. Shazeer}, \bibinfo{person}{A. Roberts}, \bibinfo{person}{K. Lee}, \bibinfo{person}{S. Narang}, \bibinfo{person}{M. Matena}, \bibinfo{person}{Y. Zhou}, \bibinfo{person}{W. Li}, {and} \bibinfo{person}{P.~J. Liu}.} \bibinfo{year}{2020}\natexlab{}.
\newblock \showarticletitle{Exploring the limits of transfer learning with a unified text-to-text transformer}.
\newblock \bibinfo{journal}{\emph{The Journal of Machine Learning Research}} \bibinfo{volume}{21}, \bibinfo{number}{1} (\bibinfo{year}{2020}), \bibinfo{pages}{5485--5551}.
\newblock


\bibitem[{Runway Research}(2024)]%
        {RunwayGen3Alpha2024}
\bibfield{author}{\bibinfo{person}{{Runway Research}}.} \bibinfo{year}{2024}\natexlab{}.
\newblock \bibinfo{title}{Introducing Gen-3 Alpha: A New Frontier for Video Generation}.
\newblock \bibinfo{howpublished}{\url{https://runwayml.com/research/introducing-gen-3-alpha}}.
\newblock


\bibitem[Saunders et~al\mbox{.}(2020)]%
        {Saunders2020ProgressiveTransformers}
\bibfield{author}{\bibinfo{person}{B. Saunders}, \bibinfo{person}{N.C. Camgöz}, {and} \bibinfo{person}{R. Bowden}.} \bibinfo{year}{2020}\natexlab{}.
\newblock \showarticletitle{Progressive Transformers for End-to-End Sign Language Production}. In \bibinfo{booktitle}{\emph{Proceedings of the European Conference on Computer Vision (ECCV)}}. \bibinfo{publisher}{Springer}, \bibinfo{address}{Cham}.
\newblock
\href{https://doi.org/10.1007/978-3-030-58600-1_43}{doi:\nolinkurl{10.1007/978-3-030-58600-1_43}}


\bibitem[Saunders et~al\mbox{.}(2021a)]%
        {Saunders2021Continuous3DSignLanguage}
\bibfield{author}{\bibinfo{person}{B. Saunders}, \bibinfo{person}{N.C. Camgöz}, {and} \bibinfo{person}{R. Bowden}.} \bibinfo{year}{2021}\natexlab{a}.
\newblock \showarticletitle{Continuous 3D Multi-Channel Sign Language Production via Progressive Transformers and Mixture Density Networks}.
\newblock \bibinfo{journal}{\emph{International Journal of Computer Vision (IJCV)}} \bibinfo{volume}{129}, \bibinfo{number}{3} (\bibinfo{year}{2021}), \bibinfo{pages}{859--877}.
\newblock
\href{https://doi.org/10.1007/s11263-020-01389-4}{doi:\nolinkurl{10.1007/s11263-020-01389-4}}


\bibitem[Saunders et~al\mbox{.}(2021b)]%
        {Saunders2021MixedSIGNals}
\bibfield{author}{\bibinfo{person}{B. Saunders}, \bibinfo{person}{N.C. Camgöz}, {and} \bibinfo{person}{R. Bowden}.} \bibinfo{year}{2021}\natexlab{b}.
\newblock \showarticletitle{Mixed SIGNals: Sign Language Production via a Mixture of Motion Primitives}. In \bibinfo{booktitle}{\emph{Proceedings of the International Conference on Computer Vision (ICCV)}}. \bibinfo{publisher}{IEEE}, \bibinfo{address}{Montreal, Canada}.
\newblock
\href{https://doi.org/10.1109/ICCV48922.2021.00456}{doi:\nolinkurl{10.1109/ICCV48922.2021.00456}}


\bibitem[Saunders et~al\mbox{.}(2021c)]%
        {Saunders2021SkeletalGraphSelfAttention}
\bibfield{author}{\bibinfo{person}{B. Saunders}, \bibinfo{person}{N.C. Camgöz}, {and} \bibinfo{person}{R. Bowden}.} \bibinfo{year}{2021}\natexlab{c}.
\newblock \showarticletitle{Skeletal Graph Self-Attention: Embedding a Skeleton Inductive Bias into Sign Language Production}.
\newblock \bibinfo{journal}{\emph{arXiv preprint arXiv:2112.05277}} (\bibinfo{year}{2021}).
\newblock
\urldef\tempurl%
\url{https://arxiv.org/abs/2112.05277}
\showURL{%
\tempurl}


\bibitem[Schembri et~al\mbox{.}(2013)]%
        {schembri2013building}
\bibfield{author}{\bibinfo{person}{A. Schembri}, \bibinfo{person}{J. Fenlon}, \bibinfo{person}{R. Rentelis}, \bibinfo{person}{S. Reynolds}, {and} \bibinfo{person}{K. Cormier}.} \bibinfo{year}{2013}\natexlab{}.
\newblock \showarticletitle{Building the British Sign Language Corpus}.
\newblock \bibinfo{journal}{\emph{Language Documentation \& Conservation}}  \bibinfo{volume}{7} (\bibinfo{year}{2013}), \bibinfo{pages}{136--154}.
\newblock


\bibitem[Secretariat(2009)]%
        {secretariat2009}
\bibfield{author}{\bibinfo{person}{United~Nations Secretariat}.} \bibinfo{year}{2009}\natexlab{}.
\newblock \bibinfo{title}{2009 Report of the United Nations Secretariat}.
\newblock
\newblock
\shownote{United Nations, New York}.


\bibitem[Segouat(2009)]%
        {Segouat2009SignLanguageCoarticulation}
\bibfield{author}{\bibinfo{person}{J. Segouat}.} \bibinfo{year}{2009}\natexlab{}.
\newblock \showarticletitle{A Study of Sign Language Coarticulation}.
\newblock \bibinfo{journal}{\emph{ACM SIGACCESS Accessibility and Computing}}  \bibinfo{volume}{94} (\bibinfo{year}{2009}), \bibinfo{pages}{41--46}.
\newblock
\urldef\tempurl%
\url{https://dl.acm.org/doi/10.1145/1597260.1597273}
\showURL{%
\tempurl}


\bibitem[Shanahan(2022)]%
        {Shanahan2022Talking}
\bibfield{author}{\bibinfo{person}{M. Shanahan}.} \bibinfo{year}{2022}\natexlab{}.
\newblock \showarticletitle{Talking about large language models}.
\newblock \bibinfo{journal}{\emph{CoRR}}  \bibinfo{volume}{abs/2212.03551} (\bibinfo{year}{2022}).
\newblock
\urldef\tempurl%
\url{https://arxiv.org/abs/2212.03551}
\showURL{%
\tempurl}


\bibitem[Taylor et~al\mbox{.}(2022)]%
        {Taylor2022Galactica}
\bibfield{author}{\bibinfo{person}{R. Taylor}, \bibinfo{person}{M. Kardas}, \bibinfo{person}{G. Cucurull}, \bibinfo{person}{T. Scialom}, \bibinfo{person}{A. Hartshorn}, \bibinfo{person}{E. Saravia}, \bibinfo{person}{A. Poulton}, \bibinfo{person}{V. Kerkez}, {and} \bibinfo{person}{R. Stojnic}.} \bibinfo{year}{2022}\natexlab{}.
\newblock \showarticletitle{Galactica: A large language model for science}.
\newblock \bibinfo{journal}{\emph{CoRR}}  \bibinfo{volume}{abs/2211.09085} (\bibinfo{year}{2022}).
\newblock
\urldef\tempurl%
\url{https://arxiv.org/abs/2211.09085}
\showURL{%
\tempurl}


\bibitem[Touvron et~al\mbox{.}(2023)]%
        {Touvron2023Llama}
\bibfield{author}{\bibinfo{person}{H. Touvron}, \bibinfo{person}{T. Lavril}, \bibinfo{person}{G. Izacard}, \bibinfo{person}{X. Martinet}, \bibinfo{person}{M. Lachaux}, \bibinfo{person}{T. Lacroix}, \bibinfo{person}{B. Rozière}, \bibinfo{person}{N. Goyal}, \bibinfo{person}{E. Hambro}, \bibinfo{person}{F. Azhar}, \bibinfo{person}{A. Rodriguez}, \bibinfo{person}{A. Joulin}, \bibinfo{person}{E. Grave}, {and} \bibinfo{person}{G. Lample}.} \bibinfo{year}{2023}\natexlab{}.
\newblock \showarticletitle{Llama: Open and efficient foundation language models}.
\newblock \bibinfo{journal}{\emph{CoRR}}  \bibinfo{volume}{abs/2302.13971} (\bibinfo{year}{2023}).
\newblock
\urldef\tempurl%
\url{https://arxiv.org/abs/2302.13971}
\showURL{%
\tempurl}


\bibitem[Wong et~al\mbox{.}(2024)]%
        {wong2024sign2gptleveraginglargelanguage}
\bibfield{author}{\bibinfo{person}{Ryan Wong}, \bibinfo{person}{Necati~Cihan Camgoz}, {and} \bibinfo{person}{Richard Bowden}.} \bibinfo{year}{2024}\natexlab{}.
\newblock \bibinfo{title}{Sign2GPT: Leveraging Large Language Models for Gloss-Free Sign Language Translation}.
\newblock
\showeprint[arxiv]{2405.04164}~[cs.CV]
\urldef\tempurl%
\url{https://arxiv.org/abs/2405.04164}
\showURL{%
\tempurl}


\bibitem[Wu et~al\mbox{.}(2024)]%
        {Li2023Unique3D}
\bibfield{author}{\bibinfo{person}{Kailu Wu}, \bibinfo{person}{Fangfu Liu}, \bibinfo{person}{Zhihan Cai}, \bibinfo{person}{Runjie Yan}, \bibinfo{person}{Hanyang Wang}, \bibinfo{person}{Yating Hu}, \bibinfo{person}{Yueqi Duan}, {and} \bibinfo{person}{Kaisheng Ma}.} \bibinfo{year}{2024}\natexlab{}.
\newblock \bibinfo{title}{Unique3D: High-Quality and Efficient 3D Mesh Generation from a Single Image}.
\newblock
\showeprint[arxiv]{2405.20343}~[cs.CV]
\urldef\tempurl%
\url{https://arxiv.org/abs/2405.20343}
\showURL{%
\tempurl}


\bibitem[Ye et~al\mbox{.}(2024)]%
        {ye2024improvingglossfreesignlanguage}
\bibfield{author}{\bibinfo{person}{Jinhui Ye}, \bibinfo{person}{Xing Wang}, \bibinfo{person}{Wenxiang Jiao}, \bibinfo{person}{Junwei Liang}, {and} \bibinfo{person}{Hui Xiong}.} \bibinfo{year}{2024}\natexlab{}.
\newblock \bibinfo{title}{Improving Gloss-free Sign Language Translation by Reducing Representation Density}.
\newblock
\showeprint[arxiv]{2405.14312}~[cs.CV]
\urldef\tempurl%
\url{https://arxiv.org/abs/2405.14312}
\showURL{%
\tempurl}


\bibitem[Yin et~al\mbox{.}(2023)]%
        {yin2023glossattentionglossfreesign}
\bibfield{author}{\bibinfo{person}{Aoxiong Yin}, \bibinfo{person}{Tianyun Zhong}, \bibinfo{person}{Li Tang}, \bibinfo{person}{Weike Jin}, \bibinfo{person}{Tao Jin}, {and} \bibinfo{person}{Zhou Zhao}.} \bibinfo{year}{2023}\natexlab{}.
\newblock \bibinfo{title}{Gloss Attention for Gloss-free Sign Language Translation}.
\newblock
\showeprint[arxiv]{2307.07361}~[cs.CV]
\urldef\tempurl%
\url{https://arxiv.org/abs/2307.07361}
\showURL{%
\tempurl}


\bibitem[Yu et~al\mbox{.}(2020)]%
        {YU2020104652}
\bibfield{author}{\bibinfo{person}{Xingxing Yu}, \bibinfo{person}{Zi-You Yu}, \bibinfo{person}{Xiao-Long Zhang}, \bibinfo{person}{Peng Li}, \bibinfo{person}{Bing Sun}, \bibinfo{person}{Xiaochun Gao}, \bibinfo{person}{Kang Yan}, \bibinfo{person}{Hao Liu}, \bibinfo{person}{Yu Duan}, \bibinfo{person}{Min-Rui Gao}, \bibinfo{person}{Guoxiu Wang}, {and} \bibinfo{person}{Shu-Hong Yu}.} \bibinfo{year}{2020}\natexlab{}.
\newblock \showarticletitle{Highly disordered cobalt oxide nanostructure induced by sulfur incorporation for efficient overall water splitting}.
\newblock \bibinfo{journal}{\emph{Nano Energy}}  \bibinfo{volume}{71} (\bibinfo{year}{2020}), \bibinfo{pages}{104652}.
\newblock
\showISSN{2211-2855}
\href{https://doi.org/10.1016/j.nanoen.2020.104652}{doi:\nolinkurl{10.1016/j.nanoen.2020.104652}}


\bibitem[Zhang et~al\mbox{.}(2023a)]%
        {zhang2023sltunetsimpleunifiedmodel}
\bibfield{author}{\bibinfo{person}{Biao Zhang}, \bibinfo{person}{Mathias Müller}, {and} \bibinfo{person}{Rico Sennrich}.} \bibinfo{year}{2023}\natexlab{a}.
\newblock \bibinfo{title}{SLTUNET: A Simple Unified Model for Sign Language Translation}.
\newblock
\showeprint[arxiv]{2305.01778}~[cs.CL]
\urldef\tempurl%
\url{https://arxiv.org/abs/2305.01778}
\showURL{%
\tempurl}


\bibitem[Zhang et~al\mbox{.}(2023b)]%
        {Zhang2023ControlNet}
\bibfield{author}{\bibinfo{person}{Lvmin Zhang}, \bibinfo{person}{Anyi Rao}, {and} \bibinfo{person}{Maneesh Agrawala}.} \bibinfo{year}{2023}\natexlab{b}.
\newblock \bibinfo{title}{Adding Conditional Control to Text-to-Image Diffusion Models}.
\newblock
\showeprint[arxiv]{2302.05543}~[cs.CV]
\urldef\tempurl%
\url{https://arxiv.org/abs/2302.05543}
\showURL{%
\tempurl}


\bibitem[Zhao et~al\mbox{.}(2021)]%
        {Zhao2021ConditionalSentenceGeneration}
\bibfield{author}{\bibinfo{person}{Jian Zhao}, \bibinfo{person}{Weizhen Qi}, \bibinfo{person}{Wengang Zhou}, \bibinfo{person}{Nan Duan}, \bibinfo{person}{Ming Zhou}, {and} \bibinfo{person}{Houqiang Li}.} \bibinfo{year}{2021}\natexlab{}.
\newblock \showarticletitle{Conditional Sentence Generation and Cross-Modal Reranking for Sign Language Translation}.
\newblock \bibinfo{journal}{\emph{IEEE Transactions on Multimedia}}  \bibinfo{volume}{24} (\bibinfo{year}{2021}), \bibinfo{pages}{2662--2672}.
\newblock
\href{https://doi.org/10.1109/TMM.2021.3074006}{doi:\nolinkurl{10.1109/TMM.2021.3074006}}


\bibitem[Zhou et~al\mbox{.}(2023)]%
        {zhou2023glossfreesignlanguagetranslation}
\bibfield{author}{\bibinfo{person}{Benjia Zhou}, \bibinfo{person}{Zhigang Chen}, \bibinfo{person}{Albert Clapés}, \bibinfo{person}{Jun Wan}, \bibinfo{person}{Yanyan Liang}, \bibinfo{person}{Sergio Escalera}, \bibinfo{person}{Zhen Lei}, {and} \bibinfo{person}{Du Zhang}.} \bibinfo{year}{2023}\natexlab{}.
\newblock \bibinfo{title}{Gloss-free Sign Language Translation: Improving from Visual-Language Pretraining}.
\newblock
\showeprint[arxiv]{2307.14768}~[cs.CV]
\urldef\tempurl%
\url{https://arxiv.org/abs/2307.14768}
\showURL{%
\tempurl}


\bibitem[Zhou et~al\mbox{.}(2021a)]%
        {Zhou2021ImprovingSignLanguageTranslation}
\bibfield{author}{\bibinfo{person}{Hao Zhou}, \bibinfo{person}{Wengang Zhou}, \bibinfo{person}{Weizhen Qi}, \bibinfo{person}{Junfu Pu}, {and} \bibinfo{person}{Houqiang Li}.} \bibinfo{year}{2021}\natexlab{a}.
\newblock \showarticletitle{Improving Sign Language Translation with Monolingual Data by Sign Back-Translation}. In \bibinfo{booktitle}{\emph{Proceedings of the IEEE/CVF Conference on Computer Vision and Pattern Recognition (CVPR)}}. \bibinfo{pages}{1316--1325}.
\newblock
\href{https://doi.org/10.1109/CVPR46437.2021.00132}{doi:\nolinkurl{10.1109/CVPR46437.2021.00132}}


\bibitem[Zhou et~al\mbox{.}(2021b)]%
        {Zhou2021SpatialTemporalMultiCueNetwork}
\bibfield{author}{\bibinfo{person}{Hao Zhou}, \bibinfo{person}{Wengang Zhou}, \bibinfo{person}{Yun Zhou}, {and} \bibinfo{person}{Houqiang Li}.} \bibinfo{year}{2021}\natexlab{b}.
\newblock \showarticletitle{Spatial-Temporal Multi-Cue Network for Sign Language Recognition and Translation}.
\newblock \bibinfo{journal}{\emph{IEEE Transactions on Multimedia}}  \bibinfo{volume}{24} (\bibinfo{year}{2021}), \bibinfo{pages}{768--779}.
\newblock
\href{https://doi.org/10.1109/TMM.2021.3054190}{doi:\nolinkurl{10.1109/TMM.2021.3054190}}


\end{thebibliography}

\appendix

\end{document}